\documentclass[runningheads]{waica}
\usepackage[T1]{fontenc}
\usepackage{graphicx}
\graphicspath{{./images/}}
\usepackage{my}
\usepackage{epigraph}
\usepackage{amssymb}
\usepackage{orcidlink}
\hypersetup{hidelinks}
\renewcommand{\orcidID}[1]{\textsuperscript{\,\raisebox{-0.05ex}{\scalebox{1.35}{\orcidlink{#1}}}}}
\begin{document}
\title{Source-Prior-Driven Selective Adaptation for Efficient Diffusion Model Finetuning}
%
\titlerunning{Selective Adaptation for Diffusion Finetuning}


\author{Yi Xiong \and
Yuan-Yuan Cheng\orcidID{0009-0005-7348-8718} \and
Xiao-Ming Fu\orcidID{0000-0001-8479-0107}}



\institute{University of Science and Technology of China, China\\
\email{1440017889@qq.com}, \email{chyy@mail.ustc.edu.cn}, \email{fuxm@ustc.edu.cn}
}


%
%
%

\maketitle              
\begin{abstract}
Fine-tuning large diffusion models for new domains or styles involves a trade-off: improving target-specific generation often degrades the pretrained model's broad generative capability. 
Existing full and parameter-efficient fine-tuning methods typically handle this trade-off only implicitly. 
In this work, we propose a novel source-prior-driven selective adaptation to efficiently finetune diffusion models, achieving a favorable trade-off.
%
Our method relies on two key observations: (1) the loss of general generative capability is highly inconsistent across pretrained parameters, and (2) those parameters that have a relatively small impact on the model's general generative capability remain structurally inconsistent across layers and parameter types.
%
Motivated by this, we 
first learn a static mask to explicitly identify parameters better suited for downstream adaptation, and then constructs structured update strategies for the selected subset. 
Experiments show that our method achieves a better adaptation-retention trade-off than the existing strong baselines.

\end{abstract}
\section{Introduction} \label{sec:introduction}

Diffusion models have become a dominant paradigm for high-quality image generation~\cite{ddpm,ddim,score_sde}. 
Latent diffusion models and their practical implementations further make large-scale text-to-image generation broadly accessible~\cite{ldm,stable_diffusion}. 
While modern pretrained diffusion models exhibit strong general-purpose generation ability, many real-world applications require adaptation to new concepts, styles, domains, or user-specific preferences~\cite{dreambooth,textual_inversion,custom_diffusion,controlnet,imagic}. 

This work focuses on finetuning with a dual objective. 
First, the adapted model should preserve as much of the pretrained model's general capability as possible. 
These capabilities constitute the main value of large pretrained diffusion models and should not be unnecessarily sacrificed during adaptation. 
Second, the model acquires strong target-specific capability for a desired domain or style. 
Thus, our aim is to achieve a trade-off between these two capabilities. 



\begin{figure}[t]
    \centering
    \begin{overpic}[width=0.93\linewidth]{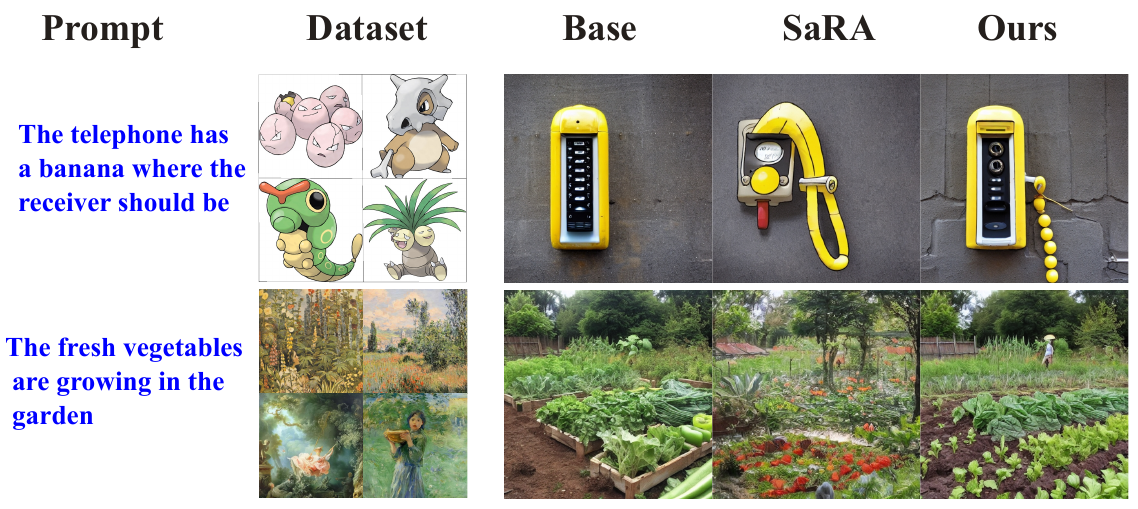}
        {
    \put(76,41.3){~\cite{sara}}
        }
    \end{overpic}
    \vspace{-3mm}
    \caption{ 
    All methods are fine-tuned on SD1.5 for the \texttt{Pokemon} and \texttt{Wikiart} domain with approximately 10M trainable parameters, and are then prompted using original COCO-style prompts rather than target-style prompts. 
    }
    \label{fig:motivation}
\end{figure}

\begin{figure}[t]
\vspace{-2mm}
    \centering
    \begin{overpic}[width=0.8\linewidth]{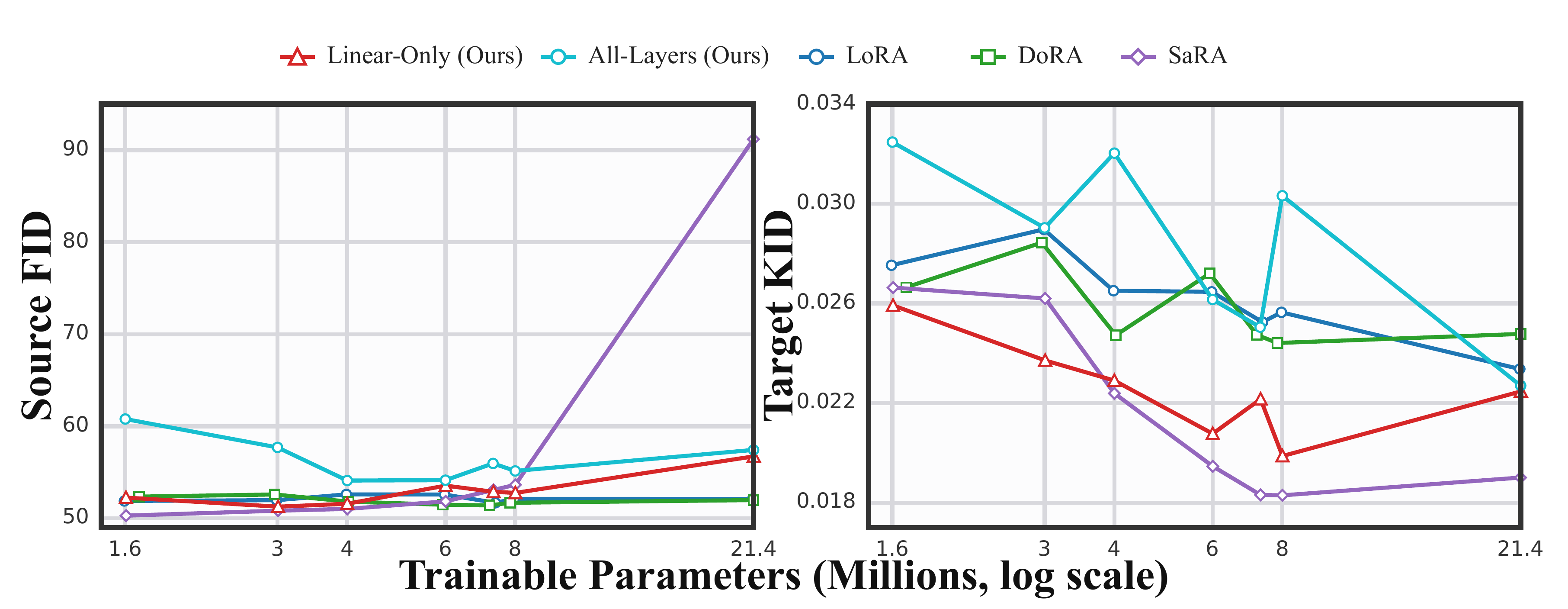}
        {
    
        }
    \end{overpic}
    \vspace{-3mm}
    \caption{Trade-off between target specialization and general capability retention 
    on SD1.5 using the \texttt{Pokemon} quick protocol. 
    }
    \vspace{-4mm}
    \label{fig:cost}
\end{figure}

Existing finetuning methods still struggle to achieve a satisfactory balance between target adaptation and capability retention (Figs.~\ref{fig:motivation} and~\ref{fig:cost}). 
Low-rank adaptation methods and knowledge-aware singular-value adaptation improve parameter efficiency by constraining the update parameterizations~\cite{lora,dora,adalora,vera,kasa}. 
Sparse or selective adaptation methods suggest that only part of the model needs to be updated for effective finetuning of diffusion models~\cite{sara,waveft,svdiff}. 
However, these methods do not explicitly optimize where new target-domain knowledge should be written so as to minimize source-side interference.

Our method is motivated by the following observation.
If parameters at certain locations are less harmful to the pretrained model's general capabilities, then prioritizing updating them for downstream adaptation may yield a better trade-off. 
This is supported by the fact that pretrained parameters are unlikely to play uniform roles in storing general knowledge and supporting specialization. 
Accordingly, we first identify a more suitable subset of parameters for finetuning.

To this end, we introduce a source-prior-driven parameter selection stage based on static mask learning to explicitly identify pretrained parameters better suited for downstream adaptation. Crucially, this source-side probing is entirely target-agnostic. It serves as a one-time offline pre-computation for a given base model and parameter budget. Once the adaptation subspace is identified, it can be cached and reused across subsequent domains or styles. 
With this subset, we further observe that different layers and parameter categories may still contribute differently to generality and specialization. 
Hence, we study these structural differences and design selective finetuning strategies over the learned mask, enabling a more controlled and effective balance. 

In summary, this paper makes two main contributions. 
\begin{enumerate}
    \item We propose a source-prior-driven parameter selection framework that discovers a target-agnostic writable subspace by mask learning. 
    \item We investigate the effects of different layers and parameter categories on the trade-off between target adaptation and general capability retention, and develop corresponding selective finetuning strategies with the learned mask.
\end{enumerate}

Extensive style adaptation experiments on both Stable Diffusion1.5~\cite{ldm} and Stable Diffusion3~\cite{pmlr-v235-esser24a} demonstrate that we achieve a better adaptation-retention balance than the baselines. 

\section{Related Work}


Besides finetuning textual embeddings, selected attention layers, or additional conditioning modules~\cite{dreambooth,textual_inversion,custom_diffusion,controlnet,imagic}, pretrained diffusion models can also be adapted to specific concepts, styles, or domains through parameter-efficient finetuning (PEFT), which updates only a small number of task-specific parameters. 



\paragraph{Low-Rank and Reparameterized PEFT.}
Classical PEFT include adapters, which insert small trainable modules into pretrained networks~\cite{adapter}, prompt- or prefix-based tuning, which optimizes continuous input-side prompts~\cite{prompt_tuning,prefix_tuning}, BitFit, which updates only bias terms~\cite{bitfit}, and Compacter, which uses low-rank hypercomplex adapter layers~\cite{compacter}. 
Among these methods, LoRA has become a standard baseline, representing weight updates as low-rank matrices~\cite{lora}. Subsequent improvements include dynamic rank allocation~\cite{adalora}, magnitude–direction decomposition~\cite{dora}, shared random projections~\cite{vera}, and timestep-dependent low-rank adaptation. 
More recent work further explores organizing, sharing, or routing low-rank updates, including knowledge-aware singular-value adaptation~\cite{kasa}, implicit rank-wise mixture-of-experts mechanism~\cite{flylora}, resource-efficient variants~\cite{relora}, centralized shadow modules~\cite{shadowpeft}, and geometry-aware or subspace-based variants~\cite{mixture_space_peft,moslora}.
These methods substantially improve the expressiveness or efficiency of PEFT by examining how to parameterize, share, or route updates. 
Unlike them, we explicitly optimize which pretrained parameter locations are safer for preserving general generative capability during domain adaptation. 


\paragraph{Sparse and Selective PEFT.}

Much recent work suggests that finetuning need not apply updates uniformly over all modules, parameters, or tokens. 
For diffusion models, SaRA observes that a portion of small-magnitude pretrained parameters can be reactivated for downstream adaptation~\cite{sara}. 
WaveFT explores sparse updates in the wavelet domain~\cite{waveft}. 
SVDiff and related compact adaptation methods also seek to reduce the update space for diffusion finetuning~\cite{svdiff}. 
These works show that sparse parameterization can be highly effective for generative model adaptation. 
More broadly, sparse training and pruning studies suggest that parameter importance is highly non-uniform across neural networks~\cite{magnitude_pruning,lottery_ticket,rigl,sparse_training_survey}.

Selective updating has been explored beyond diffusion models. 
SparseLoRA introduces contextual sparsity to reduce training compute by dynamically selecting a subset of weights for loss and gradient computation~\cite{sparselora}. 
TS-PEFT reveals token-level redundancy in standard PEFT and learns to skip PEFT updates for redundant token positions~\cite{tspeft}. 
These methods challenge dense updating but differ in granularity: TS-PEFT selects token positions, SparseLoRA enforces contextual sparsity, while we select a fixed, writable support within pretrained diffusion parameters to reduce interference with source capabilities, not just to reduce computational cost.


\paragraph{Forgetting-Aware Parameter Selection.}

Catastrophic forgetting was originally studied as the degradation of previously learned knowledge after adapting to new data~\cite{catastrophic_forgetting,ewc,lwf,continual_learning_survey}.
In continual learning, a common strategy is to estimate which parameters are important for previous tasks and then restrict or regularize their updates in the future~\cite{ewc,fisher_information,natural_gradient,kfac,lwf}. 
Other methods constrain the update space through episodic memory, parameter isolation, or mask-based importance estimation~\cite{gem,packnet,fishmask}. 
These approaches are closely related to our motivation, since they all aim to reduce interference with existing capability.

Our method differs in two aspects. 
First, instead of assigning a static importance score such as weight magnitude, gradient magnitude, or Fisher information, we learn a retention mask through source-side probing under a global writable-budget constraint. 
Second, the learned mask is not used merely as a regularizer; it is converted into a fixed update support for the subsequent target-domain finetuning stage. 
Thus, our framework directly controls where target-domain knowledge is written, providing a parameter-coordinate-level mechanism for balancing adaptation and retention in diffusion model finetuning.
\section{Method}

\subsection{Overview}

\paragraph{Problem}

Let the pretrained diffusion model be denoted by $f_{\theta}$, where $\theta$ denotes the full set of model parameters.
For notational simplicity, we treat $\theta$ as a vectorized parameter vector in the analysis, although in practice it consists of parameter tensors of different shapes.
Given target-domain data $D_t$, our goal is to adapt the pretrained model under a prescribed parameter budget while preserving as much of its general generative capability as possible. 

\paragraph{Key idea}
%

Our central hypothesis is that there exists a set of parameters that are weakly coupled to the pretrained model's general capability. 
Temporarily removing these parameters results in only a marginal increase in the original loss, yet they may still retain sufficient plasticity for downstream adaptation.
If such a low-interference set of parameters can be identified before fine-tuning and subsequent target-domain adaptation is strictly confined to that parameter set, a more favorable adaptation–retention trade-off can be achieved.

\paragraph{Methodology}


\begin{figure*}[t]
    \centering
    \includegraphics[width=0.95\linewidth]{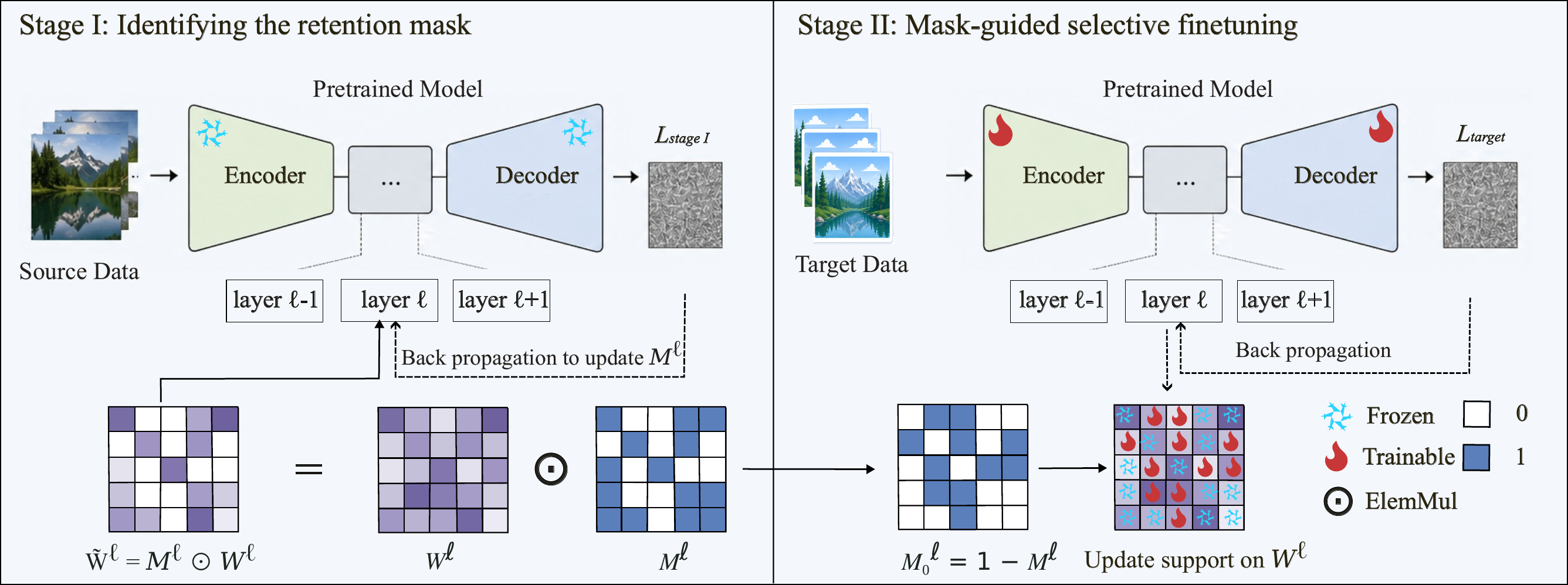}
    \vspace{-1mm}
    \caption{
    Overview of our source-prior-driven selective adaptation framework.
    }
    \label{fig:pipeline}
    \vspace{-6mm}
\end{figure*}

Our framework consists of two stages (Fig.~\ref{fig:pipeline}).
The first stage performs source-side probing and learns a \emph{retention mask} $M$, which identifies parameter coordinates whose retention best preserves source-side behavior under a writable-budget constraint (Section~\ref{sec:stage1}).
The second stage converts this retention mask into an update mask via $M_0=\mathbf{1}-M$, where $\mathbf{1}$ is the all-one vector, and performs target-domain fine-tuning only on the selected writable support (i.e., the positions of the nonzero entries of $M_0$) (Section~\ref{sec:stage2}).

\subsection{Stage I: identifying the retention mask} \label{sec:stage1}

\paragraph{Identification via parameter removal}
The objective at this stage is not to decide which positions should be updated, but rather to identify which positions cause minimal damage to the source-domain behavior upon removal, and the complement of these weakly coupled positions naturally forms the update support required for the second stage.
%
%
Thus, we learn a set of score tensors whose thresholded outputs yield a binary retention mask $M$ that indicates the parameter positions retained during source-side probing.
Then, $M_0 = \mathbf{1} - M$ denotes the positions that are actually allowed to be updated during the finetuning stage.
%

\paragraph{Source-side probed parameters.}
Given a retention mask $M$, we define the corresponding source-side probed parameter vector as $\tilde{\theta}(M) \triangleq M \odot \theta$.
Since $M_0=\mathbf{1}-M$, this can also be written as $\tilde{\theta}(M) = \theta - M_0 \odot \theta$.
This operation removes the coordinates indexed by $M_0$ from the pretrained parameter vector and evaluates the resulting source-side degradation.

\paragraph{Formulation.}
The ideal objective is to find a retention mask that minimizes the source-side probing loss $\mathcal{L}_{\text{stage1}}$ while constraining the number of writable coordinates in the complementary update mask to be less than the bound $B$:
\begin{equation}
\min_{M \in \{0,1\}^{|\theta|}} \quad \mathcal{L}_{\text{stage1}}\big(\tilde{\theta}(M)\big)
\quad \text{s.t.} \quad \lVert \mathbf{1}-M \rVert_0 \le B.
\end{equation}

\paragraph{Practical mask learning.}
Directly optimizing a binary retention mask is combinatorial. 
In practice, for each candidate weight tensor $W$ in the chosen masking scope, we introduce a learnable score tensor $S$ of the same shape.
A global threshold $\tau$ is then used to convert the scores into a hard binary retention mask:
\begin{equation}
M = \mathbf{1}[S > \tau].
\end{equation}
A high score indicates that the corresponding parameter entry is more important to source-side fidelity and should therefore be retained.

During the forward pass, the probed weight is not the original $W$, but the masked weight $W \odot M$, where $\odot$ denotes element-wise multiplication.
Since the hard thresholding operation is non-differentiable, we use a straight-through estimator (STE) for backpropagation. 
Let
$M_{\text{hard}} = \mathbf{1}[S > \tau]$ and $M_{\text{soft}} = \sigma(S)$ ($\sigma(\cdot)$ is the sigmoid function), 
the effective mask used in training is
\begin{equation}
\hat{M} = M_{\text{hard}}.\mathrm{detach}() - M_{\text{soft}}.\mathrm{detach}() + M_{\text{soft}},
\end{equation}
where $\operatorname{detach}(\cdot)$ is the stop-gradient operator.
The forward pass uses the hard binary mask, and the backward pass follows the gradient of the soft relaxation.

\paragraph{Score-learning objective.}
The score tensors are optimized on source-domain data so that the masked model preserves source-side behavior while maintaining the desired sparsity level. 
Concretely, the Stage I objective is
\begin{equation}
\mathcal{L}_{\text{stage1}} = \mathcal{L}_{\text{src}} + \lambda \sum_i \sigma(S_i).
\end{equation}
where $\mathcal{L}_{\text{src}}$ denotes the source-side fidelity loss and the second term is a sparsity regularizer over all score entries.
For example, for SD1.5, $\mathcal{L}_{\text{src}}$ is the standard diffusion noise-prediction MSE loss, and for SD3, $\mathcal{L}_{\text{src}}$ is replaced by the corresponding flow-matching objective. 
In both cases, the regularizer $\sum_i \sigma(S_i)$ encourages more score entries to decrease, so that fewer positions are retained and more positions become writable in the complementary update mask.

The threshold $\tau$ is not fixed throughout training. 
Instead, every few iterations, we aggregate all score entries across the candidate parameter space and recompute a global threshold using k-th value selection, ensuring the resulting masked-out ratio matches the prescribed writable budget. 
After Stage I converges, the final writable mask is exported as $M_0 = \mathbf{1}[S \le \tau]$,
while the final retention mask is $M = \mathbf{1}[S > \tau].$

\paragraph{Second-order interpretation.}
Although Stage I is implemented through score tensors and threshold-based masking, its behavior can be interpreted through the following discrete support-selection view.
To gain deeper insight into the probing objective, we perform a second-order Taylor expansion of $\mathcal{L}_{\text{stage1}}(\theta - M_0 \odot \theta)$ around the pretrained parameters $\theta$. Let $D_{M_0}$ be the diagonal selection matrix induced by the vectorized mask $M_0$. We then have:
\begin{equation}
\begin{aligned}
\mathcal{L}_{\text{stage1}}(\theta - M_0 \odot \theta) - \mathcal{L}_{\text{stage1}}(\theta)
\approx &- g_s^\top (M_0 \odot \theta) + \frac{1}{2} (M_0 \odot \theta)^\top H_s (M_0 \odot \theta) \\
= &- g_s^\top D_{M_0} \theta + \frac{1}{2} \theta^\top D_{M_0} H_s D_{M_0} \theta,
\end{aligned}
\end{equation}
where $g_s = \nabla_{\theta} \mathcal{L}_{\text{stage1}}(\theta)$ is the source-domain gradient and $H_s$ is the Hessian matrix. 
If $\theta$ lies in the neighborhood of a (local) optimum on the source domain, the first-order gradient term $g_s$ is typically small. 
Therefore, minimizing the probing loss is approximately equivalent to minimizing the second-order term $\frac{1}{2}\theta^\top D_{M_0} H_s D_{M_0} \theta$.
This shows that Stage I tends to identify parameter coordinates whose removal incurs only a small source-side quadratic penalty.

\subsection{Stage II: Mask-guided selective finetuning}\label{sec:stage2}

\paragraph{Constrained optimization.}

After Stage I, we fix the learned update mask $M_0$, which serves as the sole support for parameter updates in Stage II. 
The target-domain finetuning is formulated as a constrained optimization problem:
\begin{equation}
\min_{u} \quad \mathcal{L}_\text{target}\big(\theta + D_{M_0}u\big),
\end{equation}
where $\mathcal{L}_\text{target}$ is the target-domain loss, $D_{M_0}$ is the diagonal selection matrix induced by $M_0$, and $u$ denotes the free update coefficients on the selected support.
Equivalently, the actual parameter update can be written as $\Delta\theta = D_{M_0}u$.
Thus, all target-domain gradients can be injected into the model only through the coordinates marked by $M_0$, while all remaining parameters stay frozen.

\paragraph{A restricted Hessian perspective.}

From the view of preserving pretrained capability, the source-side loss change induced by the second-stage update satisfies: 
\begin{equation}
\begin{aligned}
\mathcal{L}_{\text{stage1}}(\theta+\Delta\theta) - \mathcal{L}_{\text{stage1}}(\theta)
&\approx g_s^\top \Delta\theta + \frac{1}{2} \Delta\theta^\top H_s \Delta\theta \\
&= g_s^\top D_{M_0} u + \frac{1}{2} u^\top D_{M_0} H_s D_{M_0} u.
\end{aligned}
\end{equation}
Near the pretrained solution, the gradient term $g_s$ may be small, and the source loss increment may be dominated by the quadratic form of the restricted Hessian $D_{M_0} H_S D_{M_0}$. 
Since Stage I has already biased $M_0$ toward low-interference parameters, this quadratic form tends to remain small during the update, thereby preserving general generation capability. 

\paragraph{Adaptation capacity.}
For target-domain fine-tuning to remain effective, the projected target-domain gradient $D_{M_0}g_t$ (with $g_t=\nabla_{\theta}\mathcal{L}_\text{target}$) should not be too small.
This motivates our design principle: the selected support should not only be low-interference with respect to pretrained capability, but also sufficiently expressive for downstream fine-tuning.
Under this view, effective adaptation is achieved not by updating the entire parameter space, but by writing target-domain knowledge into a restricted yet still trainable support.

\begin{figure}[t]
    \centering
     \includegraphics[width=0.6\linewidth]{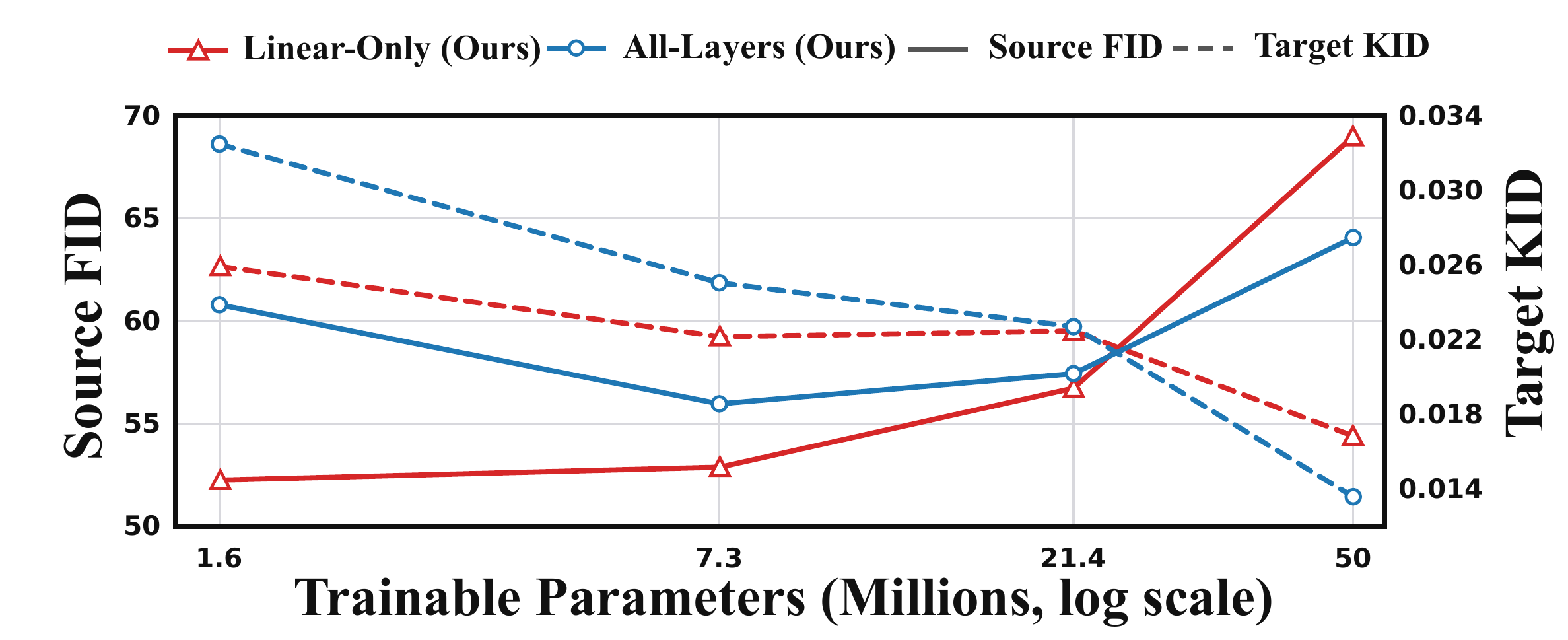}
     \vspace{-3mm}
    \caption{\emph{Linear-only} vs. \emph{All-layers} under different parameter budgets.}
     \label{fig:ablation-scope}
\end{figure}

\subsection{Mask Selection for Diffusion Model Finetuning}


We instantiate the above framework into two variants to accommodate different layer types in the denoising network of the diffusion model:
\begin{itemize}
    \item \emph{Linear-only}: the mask is applied solely to the linear layers within attention modules and feed-forward networks;
    \item \emph{All-layers}: the mask is applied to a broader set of all layers.
\end{itemize}


Our experiments indicate that under extremely small parameter update budgets, the \emph{Linear-only} variant often outperforms the All-layers variant with the same budget (Fig.~\ref{fig:ablation-scope}). 
When the budget is very small, allowing selection from all linear and convolutional layers greatly expands the search space, causing the support to disperse over many coordinates that are not critical for fine-tuning and making it difficult to form a concentrated and effective adaptation channel. 

\section{Experiments and Results}\label{sec:results}

\begin{table*}[!t]
\centering
\scriptsize
\setlength{\tabcolsep}{4pt}
\caption{
Comparisons on SD1.5 and SD3.
All target/source metrics are reported using the (5k target/50k
source) protocol, where available. 
The Trade-off is computed within each dataset and each parameter group.}
\label{tab:sd15_main_results_grouped}
\resizebox{0.765\textwidth}{!}{
\begin{tabular}{llccccc}
\toprule
\multicolumn{7}{c}{\textbf{WikiArt (SD1.5)}} \\
\midrule
Params & Method & tKID$(\times 10^3)\downarrow$ & tCLIP$(\times 10^2)\uparrow$ & sFID$\downarrow$ & sCLIP$(\times 10^2)\uparrow$ & Trade-off$\downarrow$ \\
\midrule
\multirow{4}{*}{1.6M} & LoRA (r=8) & 47.62 & 28.67 & 18.71 & 31.46 & 1.07 \\
& DoRA (r=8) & 46.33 & 28.61 & 18.85 & 31.47 & 1.05 \\
& SaRA & 54.61 & \textbf{28.94} & \textbf{18.48} & 31.46 & 1.21 \\
& Linear-Only & \textbf{45.03} & 28.59 & 19.20 & \textbf{31.50} & \textbf{1.04} \\
\cmidrule(lr){1-7}
\multirow{5}{*}{21.4M} & LoRA & 40.45 & \textbf{28.08} & 18.93 & \textbf{31.39} & 2.41 \\
& DoRA & 39.46 & 28.03 & 18.70 & 31.28 & 2.32 \\
& SaRA & 25.59 & 27.28 & 33.16 & 29.36 & 2.67 \\
& Linear-Only & \textbf{17.15} & 26.86 & \textbf{18.52} & 31.25 & \textbf{1.00} \\
& All-Layers & 30.81 & 27.63 & 19.13 & 31.38 & 1.86 \\
\cmidrule(lr){1-7}
\multirow{4}{*}{50M} & LoRA & 32.36 & 27.76 & \textbf{18.65} & 30.85 & 1.84 \\
& DoRA & 33.08 & \textbf{27.78} & 18.92 & 30.70 & 1.91 \\
& SaRA & \textbf{17.55} & 27.27 & 55.71 & 27.12 & 2.99 \\
& All-Layers & 22.24 & 27.12 & 22.81 & \textbf{30.98} & \textbf{1.55} \\
\midrule
\multicolumn{7}{c}{\textbf{Anime (SD1.5)}} \\
\midrule
Params & Method & tKID$(\times 10^3)\downarrow$ & tCLIP$(\times 10^2)\uparrow$ & sFID$\downarrow$ & sCLIP$(\times 10^2)\uparrow$ & Trade-off$\downarrow$ \\
\midrule
\multirow{4}{*}{1.6M} & LoRA (r=8) & 63.44 & \textbf{24.36} & 19.94 & 31.44 & 1.76 \\
& DoRA (r=8) & \textbf{36.59} & 24.01 & \textbf{19.61} & 31.47 & \textbf{1.00} \\
& SaRA & 41.44 & 23.32 & 27.44 & 31.30 & 1.58 \\
& Linear-Only & 39.47 & 23.46 & 21.71 & \textbf{31.47} & 1.19 \\
\cmidrule(lr){1-7}
\multirow{5}{*}{21.4M} & LoRA & 30.01 & 23.35 & \textbf{20.64} & 31.37 & 1.35 \\
& DoRA & 26.23 & 23.62 & 21.00 & 31.37 & \textbf{1.20} \\
& SaRA & 30.41 & 23.50 & 90.70 & 28.24 & 6.00 \\
& Linear-Only & \textbf{22.28} & \textbf{23.87} & 33.11 & 31.35 & 1.60 \\
& All-Layers & 29.73 & 23.52 & 35.51 & \textbf{31.45} & 2.30 \\
\cmidrule(lr){1-7}
\multirow{4}{*}{50M} & LoRA & 34.90 & 23.84 & \textbf{21.68} & 31.32 & 1.53 \\
& DoRA & 24.20 & 23.32 & 21.87 & \textbf{31.40} & \textbf{1.07} \\
& SaRA & 38.06 & \textbf{24.24} & 180.93 & 22.19 & 13.95 \\
& All-Layers & \textbf{22.77} & 23.61 & 87.20 & 30.70 & 4.02 \\
\midrule
\multicolumn{7}{c}{\textbf{Cyberpunk (SD1.5)}} \\
\midrule
Params & Method & tKID$(\times 10^3)\downarrow$ & tCLIP$(\times 10^2)\uparrow$ & sFID$\downarrow$ & sCLIP$(\times 10^2)\uparrow$ & Trade-off$\downarrow$ \\
\midrule
\multirow{4}{*}{1.6M} & LoRA (r=8) & 10.98 & 35.34 & \textbf{20.97} & 31.48 & 1.49 \\
& DoRA (r=8) & 11.22 & 35.41 & 21.53 & \textbf{31.51} & 1.56 \\
& SaRA & 11.22 & 35.59 & 23.26 & 31.29 & 1.69 \\
& Linear-Only & \textbf{7.37} & \textbf{35.83} & 22.00 & 31.36 & \textbf{1.05} \\
\cmidrule(lr){1-7}
\multirow{5}{*}{21.4M} & LoRA & 10.23 & 35.74 & 22.40 & 31.43 & 1.41 \\
& DoRA & 9.38 & 35.88 & \textbf{21.94} & \textbf{31.55} & 1.27 \\
& SaRA & 10.29 & 35.73 & 30.65 & 30.72 & 1.95 \\
& Linear-Only & \textbf{7.39} & \textbf{36.42} & 26.17 & 31.28 & \textbf{1.19} \\
& All-Layers & 8.71 & 36.33 & 25.43 & 31.29 & 1.37 \\
\cmidrule(lr){1-7}
\multirow{4}{*}{50M} & LoRA & 9.76 & 35.90 & \textbf{22.56} & 31.38 & 1.04 \\
& DoRA & \textbf{9.40} & 35.97 & 22.59 & \textbf{31.40} & \textbf{1.00} \\
& SaRA & 10.71 & 35.67 & 40.75 & 29.93 & 2.06 \\
& All-Layers & 10.25 & \textbf{36.24} & 30.00 & 31.01 & 1.45 \\
\midrule
\multicolumn{7}{c}{\textbf{Pokemon (SD3)}} \\
\midrule
Params & Method & tKID$(\times 10^3)\downarrow$ & tCLIP$(\times 10^2)\uparrow$ & sFID$\downarrow$ & sCLIP$(\times 10^2)\uparrow$ & Trade-off$\downarrow$ \\
\midrule
\multirow{5}{*}{2.5M} & LoRA (r=4) & 17.33 & 33.08 & 26.97 & 32.05 & 1.22 \\
& DoRA (r=4) & 18.53 & 33.09 & 27.98 & \textbf{32.10} & 1.35 \\
& SaRA & 18.25 & \textbf{33.52} & \textbf{24.58} & 31.95 & 1.17 \\
& Linear-Only & 17.24 & 33.34 & 24.77 & 31.92 & 1.12 \\
& All-Layers & \textbf{15.56} & 33.47 & 25.03 & 31.96 & \textbf{1.02} \\
\cmidrule(lr){1-7}
\multirow{5}{*}{5M} & LoRA (r=8) & 18.87 & 33.13 & 28.87 & 32.07 & 1.55 \\
& DoRA (r=8) & 20.39 & 33.24 & 28.59 & \textbf{32.15} & 1.66 \\
& SaRA & 20.19 & 33.12 & 32.26 & 31.78 & 1.85 \\
& Linear-Only & 16.30 & 33.20 & \textbf{24.65} & 31.98 & 1.14 \\
& All-Layers & \textbf{14.29} & \textbf{33.25} & 24.75 & 31.98 & \textbf{1.00} \\
\cmidrule(lr){1-7}
\multirow{5}{*}{10M} & LoRA (r=16) & 21.17 & \textbf{33.32} & 27.77 & 32.06 & 1.52 \\
& DoRA (r=16) & 21.71 & 33.27 & 28.29 & \textbf{32.06} & 1.59 \\
& SaRA & 48.97 & 32.48 & 47.69 & 30.66 & 6.03 \\
& Linear-Only & 19.58 & 33.17 & \textbf{24.78} & 32.00 & 1.25 \\
& All-Layers & \textbf{15.63} & 33.12 & 25.31 & 32.00 & \textbf{1.02} \\
\bottomrule
\end{tabular}}
\vspace{-7mm}
\end{table*}

\paragraph{Experimental Setup.}
We conduct experiments on two diffusion backbones with distinct target-domain datasets: \texttt{WikiArt}, \texttt{Cyberpunk}, and \texttt{Anime} for SD1.5, and \texttt{Pokemon} for SD3.
We compare against three representative parameter-efficient fine-tuning baselines, LoRA~\cite{lora}, DoRA~\cite{dora}, and SaRA~\cite{sara} that together span the most widely used low-rank and sparse adaptation strategies for diffusion models.
We evaluate target-domain adaptation with Kernel Inception Distance (KID)~\cite{kid} and CLIP score~\cite{clip}, and source-domain retention with Fr\'echet Inception Distance (FID)~\cite{fid} and CLIP score, and 
summarize their balance via a Trade-off score, defined within each dataset-budget group as $\text{Trade\text{-}off} = (\mathrm{tKID}/\min(\mathrm{tKID})) \times (\mathrm{sFID}/\min(\mathrm{sFID}))$ (lower is better).
To balance evaluation fidelity and cost, we use two protocols that share identical training configurations but differ only in the number of generated samples: (1) a \emph{quick protocol} (1k target/1k source images) used for broad comparison and ablations, and (2) a \emph{rigorous protocol} (5k target/50k source images) for careful comparison. 
Full details of metric choices, sample sizes, and parameter budgets are deferred to the supplementary material. 

\paragraph{Comparisons in rigorous protocol.}
As shown in Table~\ref{tab:sd15_main_results_grouped}, the rigorous protocol shows that the best adaptation--retention trade-off depends on both the target domain and the parameter budget. At the low-budget setting (1.6M), all methods are relatively close: \emph{Linear-Only} performs best on \texttt{WikiArt} and \texttt{Cyberpunk}, while DoRA slightly leads on \texttt{Anime}.

At the medium-budget setting (21.4M), our \emph{Linear-Only} variant achieves the best Trade-off on \texttt{WikiArt} and \texttt{Cyberpunk}, and also gives the lowest tKID on these datasets. On \texttt{Anime}, DoRA obtains a better Trade-off due to stronger source-domain retention, although \emph{Linear-Only} remains competitive on target-side adaptation.

At the large-budget setting (50M), the results become more mixed: \emph{All-Layers} gives the best Trade-off on \texttt{WikiArt}, while DoRA performs best on \texttt{Anime} and \texttt{Cyberpunk}. SaRA often improves target adaptation but can substantially degrade source-domain fidelity, leading to worse Trade-off scores. On SD3 \texttt{Pokemon}, \emph{All-Layers} consistently achieves the best Trade-off across all evaluated budgets, indicating that a broader writable parameter space is more effective for the larger backbone.

\begin{figure*}[t]
    \centering
    \begin{overpic}[width=0.98\textwidth]{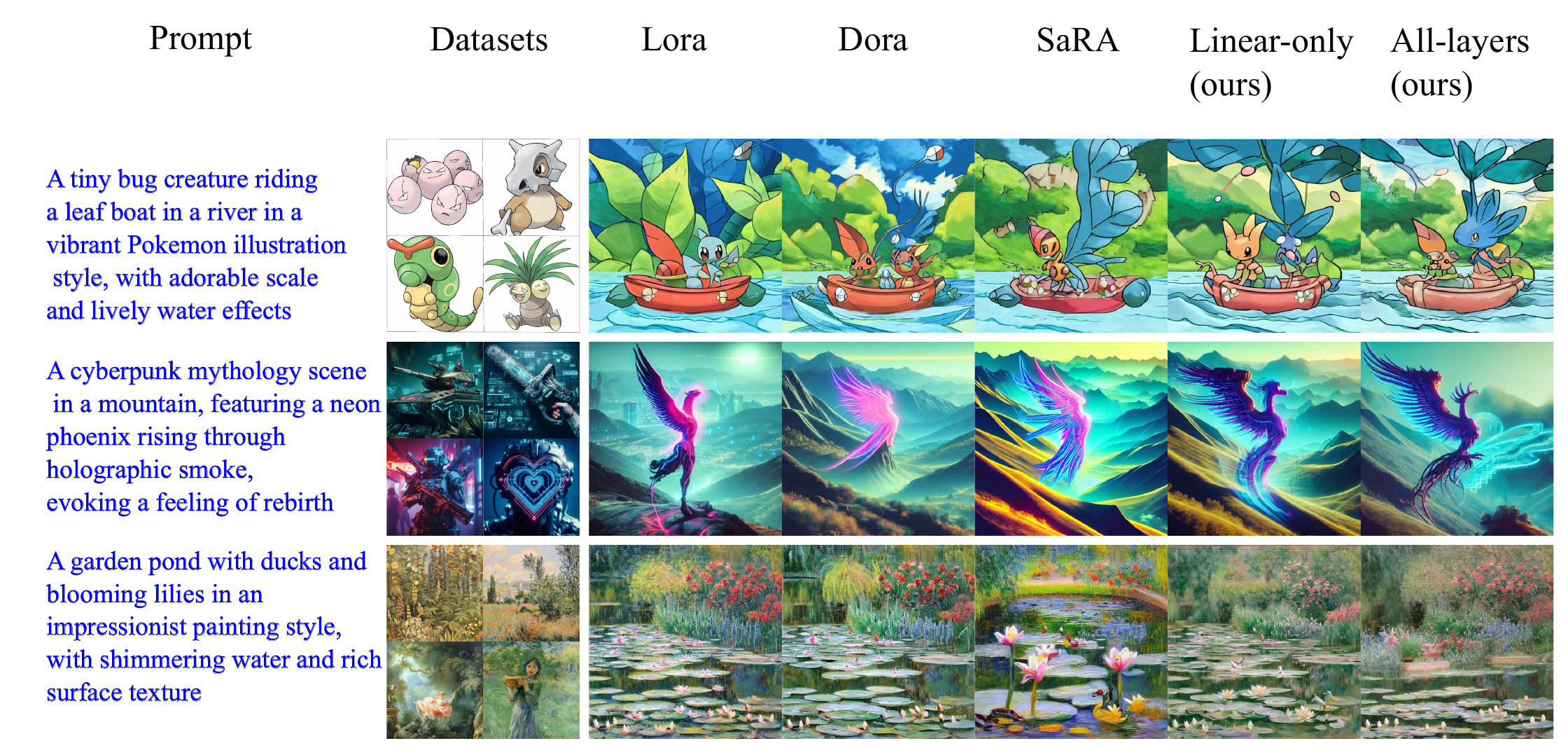}
    \end{overpic}
    \vspace{-3mm}
    \caption{Comparisons of different fine-tuning methods on representative prompts. 
    }
    \label{fig:comparison1}
\end{figure*}

\paragraph{Dense Low-Budget Comparisons.}
To complement the sparse rigorous evaluation, we conduct a denser low-budget sweep under the quick protocol on \texttt{Pokemon}, \texttt{Cyberpunk}, and \texttt{Anime} at the 3M, 4M, 6M, and 8M budgets (see Table 2 in the supplementary material).
On \texttt{Pokemon} and \texttt{Cyberpunk}, \emph{Linear-only} attains the best trade-off at every budget. 
The improvement is driven primarily by a substantially lower KID: at 8M, our tKID of 19.87/7.240 is 18.6\% on Pokemon and 22.6\% on Cyberpunk lower than the best baseline, while sFID increases by only $\sim$2--4\%. 
On \texttt{Anime}, the picture is more mixed: \emph{Linear-Only} leads at 3M, but DoRA catches up at 4M and 8M. 
This shows that for highly stylized, distribution-shifted target domains, the marginal benefit of selective writing diminishes once the budget is sufficient for low-rank baselines to fit the target distribution.

Taken together with the rigorous results, these denser sweeps confirm that the gain of our method is most pronounced in the small-to-medium budgets, where every writable parameter must be allocated appropriately.
Moreover, we provide qualitative comparisons in Fig.~\ref{fig:comparison1}, which shows the superior generation quality of our method.  

\paragraph{Additional Compact and Selective Baselines.}
We further evaluate two compact or selective diffusion fine-tuning methods, SVDiff and WaveFT, with full results provided in Appendix.
SVDiff is evaluated under its original singular-value-only parameterization, which results in an ultra-compact setting with about 0.2M trainable parameters in our SD1.5 implementation.
It preserves source-domain capability well, but does not naturally support the budget-controlled comparison used in our main experiments.
WaveFT is additionally tested under a matched 1.6M trainable-parameter budget.
The results show that directly scaling WaveFT to this budget causes substantial source-side degradation, with much higher source FID than our 1.6M Linear-Only variant.
These comparisons suggest that SVDiff and WaveFT are informative compact-adaptation baselines, while our method is more aligned with the budget-controlled adaptation--retention trade-off studied in this work.

\paragraph{Human Preference Study.}
To complement the automatic FID/KID/CLIP-based evaluation, we further conduct a pilot blinded pairwise human preference study.
Each trial presents two anonymized images generated from the same prompt by our method and one baseline, and participants choose between ``A is better'', ``B is better'', or ``Tie / No clear preference''.
Method names and metric scores are hidden, and the left--right order is randomized.
We evaluate two aspects aligned with our objective: target-domain adaptation and source-domain retention.

For target adaptation, our method achieves an average tie-adjusted preference of 70.8\% across domain--baseline comparison groups, indicating that human raters generally prefer the target-domain images produced by our method.
For source retention, the average tie-adjusted preference is 52.1\%, showing a more mixed pattern: our method is consistently preferred over SaRA, but not uniformly over LoRA or DoRA.
These results are consistent with our quantitative findings: the proposed method improves the adaptation--retention trade-off in many budget-constrained regimes, but is not a universal winner on every individual target or source criterion.
Detailed protocols and complete preference statistics are provided in Appendix.


\subsection{Ablation Study}\label{sec:ablation}

Unless otherwise specified, all ablations follow the same training and evaluation pipeline as the main experiments to isolate the contribution of each factor.

\begin{figure}[t]
    \centering
    \includegraphics[width=0.99\linewidth]{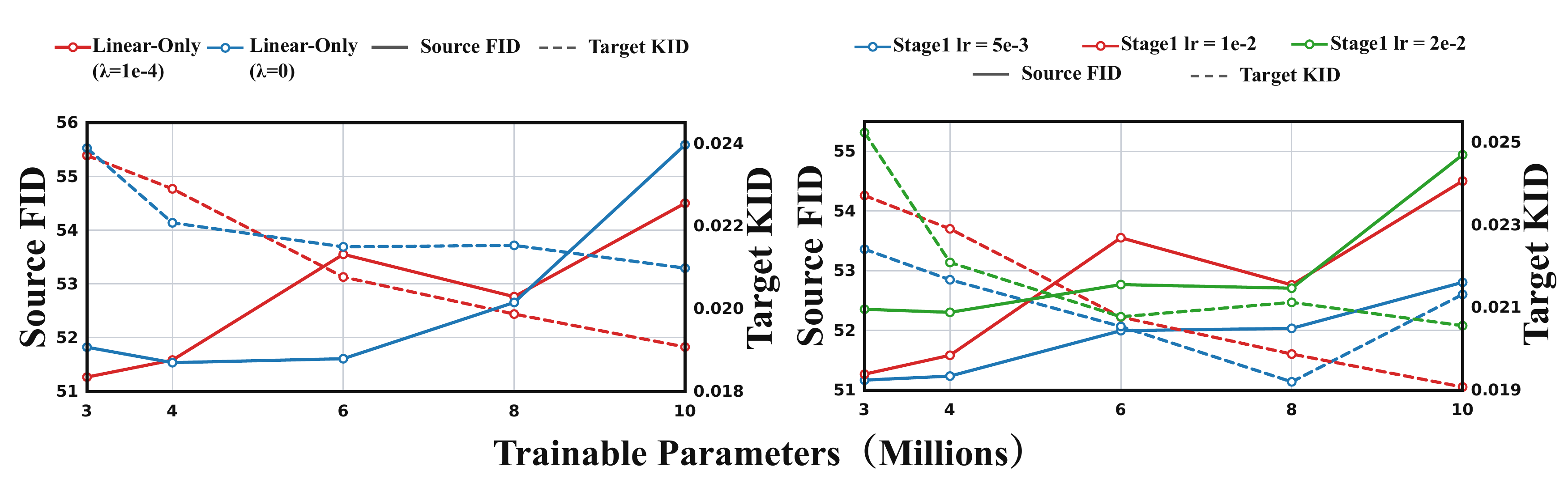}
    \vspace{-4mm}
    \caption{
    Left: Ablation on sparse loss, comparing $\lambda=1e-4$ and $\lambda=0$ under different parameter budgets. 
    Right: Ablation on learning rates.
    }
    \label{fig:ablation-loss-lr}
\end{figure}

\paragraph{Effect of the Sparsity Loss.}
Stage I optimizes the source-side fidelity objective with a sparsity penalty. We compare the full method against a variant without the sparsity regularizer (Fig.~\ref{fig:ablation-loss-lr} - Left) to test whether it is necessary for inducing a meaningful writable support, rather than relying on thresholding alone.

\paragraph{Learning Rate Sensitivity.}
Because our method restricts adaptation to a selected writable support, the effective optimization dynamics may differ from standard PEFT baselines. Fig.~\ref{fig:ablation-loss-lr} - Right examines the sensitivity to the target-domain learning rate, showing that the observed performance gain is robust across a reasonable range rather than relying on a narrowly tuned setting.

\begin{table}[t]
    \centering
    \caption{
    Ablation on learned vs. random masks under the same writable budget.
    }
    \label{tab:ablation-random}
    \resizebox{0.7\linewidth}{!}{
    \begin{tabular}{l c c c c}
        \toprule
        Budget & Method & Target KID$(\times 10^3)\downarrow$ & Source KID$(\times 10^3)\downarrow$ & Source KID std$(\times 10^3)\downarrow$ \\
        \midrule
        \multirow{3}{*}{12M} & Linear random & 20.94 & 6.09 & 4.52 \\
        & Global random & 22.90 & 6.01 & 4.46 \\
        & Linear-only (Ours) & \textbf{20.06} & \textbf{5.67} & \textbf{3.72} \\
        \bottomrule
    \end{tabular}
    }
\end{table}

\paragraph{Learned Mask vs. Random Mask.}
To verify that the gain does not come merely from restricting updates to a sparse subset, we compare the learned mask against random masks under the same writable budget (Table~\ref{tab:ablation-random}). The learned mask achieves the best target KID, source KID, and source KID variance, indicating that source-prior-driven selection identifies a more effective writable support than unstructured sparse selection of equal size.


\subsection{Analyzing Learned Mask Distribution}
\label{sec:mask-analysis}

\begin{figure}[t]
    \centering
    \begin{overpic}[width=0.75\linewidth]{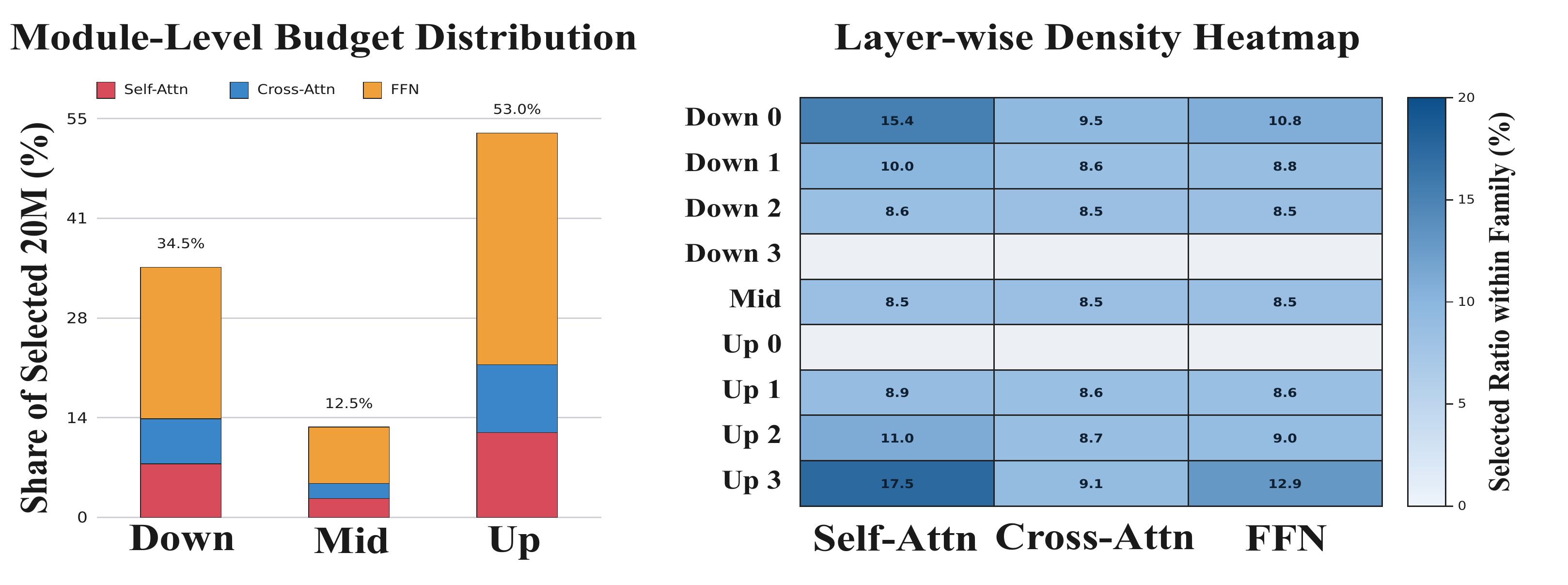}
    \end{overpic}
    \vspace{-4mm}
    \caption{Visualization of the learned mask distribution across layers (SD1.5, 20M selected parameters). The selected update positions are allocated in a structured, non-uniform manner across modules.}
    \label{fig:mask_distribution}
\end{figure}

To better understand \emph{why} the proposed selection is effective, we visualize the distribution of the writable support discovered by Stage I across the diffusion model's modules and layer families. Fig.~\ref{fig:mask_distribution} reports both a module-level budget summary and a layer-wise density heatmap under the SD1.5 setting with a 20M-parameter budget.

The learned support is clearly non-uniform. The budget is distributed unevenly across the down, mid, and up blocks of the network. Within each layer family, the density varies across self-attention, cross-attention, and feed-forward submodules. 
This structured allocation pattern is consistent with our main design intuition that effective target-domain fine-tuning benefits from writing new knowledge into a selectively chosen subspace, rather than spreading updates uniformly over the entire model.

\section{Conclusion and Discussion}\label{sec:conclusion}
We studied the adaptation–retention trade-off in diffusion model finetuning, focusing on where to write target-domain knowledge to reduce interference with  the pretrained generative capability. 
We proposed a source-prior-driven selective adaptation framework that learns a static retention mask via source-side probing and uses its complement as a fixed, writable support, thereby restricting updates to coordinates less coupled to the source domain. 
Experiments on SD1.5 and SD3 show the method achieves competitive target-domain performance while better preserving open-domain quality, especially under limited trainable budgets. 
Mask visualization reveals that the learned support is structured and non-uniform, unlike that of random sparse subsets.

There are several limitations. 
First, the quality of the learned mask depends on the source-side probing distribution, and a more diverse or better-calibrated source set may further improve retention. 
Second, although Stage I is performed only once, it still introduces additional computational cost.
Third, the current framework learns a static parameter-coordinate support, while different prompts, timesteps, or target domains may benefit from more dynamic update routing. 
Future work could extend the proposed source-prior-driven selection to prompt-aware, timestep-aware, or task-adaptive masks and explore its applicability to broader generative backbones and multi-domain continual adaptation scenarios.
\section*{Acknowledgments}\label{sec:acknowledgments}
The authors would like to thank the anonymous reviewers for their constructive suggestions and comments.
This work is supported by the Fundamental and Interdisciplinary Disciplines Breakthrough Plan of the Ministry of Education of China and the National Natural Science Foundation of China (92570201, 62272429).
\clearpage
\appendix
\section{Detailed Experimental Setup}
\label{app:experimental-setup}

This appendix provides the implementation details that are summarized in the main paper's experimental setup.

\paragraph{Evaluation perspectives and metrics.}
All methods are evaluated from two complementary perspectives: target-domain adaptation and source-domain capability retention.
We employ three standard metrics:
(i) \emph{Fr\'echet Inception Distance} (FID)~\cite{fid} measures the distance between two image distributions by fitting Gaussians to Inception-V3 features and computing their Fr\'echet distance; lower values indicate better distribution matching.
(ii) \emph{Kernel Inception Distance} (KID)~\cite{kid} computes the squared maximum mean discrepancy (MMD) with a polynomial kernel over Inception features. Unlike FID, KID is an unbiased estimator and is more reliable at smaller sample sizes; lower is better.
(iii) \emph{CLIP score}~\cite{clip} measures the cosine similarity between CLIP image and text embeddings, reflecting semantic alignment between generated images and prompts; higher is better.
For target-domain adaptation we report KID (denoted tKID) and CLIP (tCLIP), and for source-domain retention we primarily report FID (sFID) and CLIP (sCLIP). We use KID as an unbiased estimator of the FID-style distribution discrepancy that is better suited to relatively small sample sizes, and reserve the more expensive FID for the rigorous evaluation.

\paragraph{Quick protocol.}
The quick protocol is used for most experiments and ablation studies to enable efficient iteration. We generate 1{,}000 images for the target domain and 1{,}000 images for the source domain. For broad multi-setting comparisons we report FID on the source side, while a small number of lightweight diagnostic experiments use KID instead for efficiency.

\paragraph{Rigorous protocol.}
The rigorous protocol is applied only to a few representative parameter settings in the main result tables. We generate 5{,}000 target-domain images and report KID, and 50{,}000 source-domain images and report FID. The two protocols differ only in the number of generated samples; all training configurations remain identical. Evaluating every setting with the 5{,}000/50{,}000-sample configuration would be prohibitively expensive, so we use the quick protocol for broad comparison and hyperparameter exploration and reserve the rigorous protocol for selected representative settings.

\paragraph{Trade-off metric.}
To summarize the balance between adaptation and retention with a single number, we report a \texttt{Trade-off} score. Within each dataset--budget group, we normalize the target and source metrics by their respective minima across all compared methods,
\begin{equation}
\texttt{Knorm} = \frac{tKID}{\min(tKID)},
\qquad
\texttt{Fnorm} = \frac{sFID}{\min(sFID)},
\end{equation}
and define
\begin{equation}
\texttt{Trade\text{-}off} = \texttt{Knorm} \times \texttt{Fnorm}.
\end{equation}
A lower \texttt{Trade-off} indicates a better joint balance between target-domain adaptation and source-domain capability retention.
\paragraph{Choice of parameter budgets for the rigorous protocol.}
Because the rigorous protocol requires substantially more generated samples, we restrict the main rigorous comparison to a small set of representative parameter budgets. For LoRA and DoRA we adopt the commonly used rank-$8$ configuration as the standard low-budget setting, which corresponds to approximately 1.6M trainable parameters in our SD1.5 implementation. For higher-budget comparisons, we additionally include the 20M and 50M settings that are widely used in prior sparse adaptation benchmarks. This sparse rigorous evaluation is complemented by denser budget sweeps under the quick protocol, in which a broader range of LoRA/DoRA configurations is reported.

\section{Computational Cost}
All experiments are conducted on three NVIDIA RTX 4090 GPUs. 
For the rigorous evaluation protocol, running one method under one configuration takes approximately 20 GPU-hours on SD1.5 and approximately 5 GPU-days on SD3. 
The substantially higher cost of SD3 mainly comes from the larger backbone and the more expensive generation process under the 5k/50k-sample evaluation setting.

In total, completing all training, evaluation, and ablation experiments took approximately one and a half months of wall-clock time using the available hardware. 
This computational cost motivates our use of two evaluation protocols: the quick protocol is used for broad budget sweeps and ablation studies, while the rigorous protocol is reserved for representative configurations in the main comparison.

\section{Empirical Validation of the Taylor Expansion Assumption}
\label{app:empirical_validation}

In Section 3.2 and Section 3.3 of the main paper, our theoretical motivation relies on a second-order Taylor expansion (Eq. 5 and Eq. 7), which assumes that the pretrained parameters $\theta$ reside near a local optimum of the source domain. Consequently, the first-order gradient term $g_s^\top \Delta\theta$ is assumed to be sufficiently small, allowing the second-order quadratic form $\frac{1}{2}\Delta\theta^\top H_s \Delta\theta$ to dominate the source-loss variation. 

To empirically validate this assumption without explicitly materializing the intractable Hessian matrix $H_s$, we track the absolute magnitudes of both the first-order and second-order terms during the Stage I source-side probing process. The quadratic term is efficiently computed using the Hessian-Vector Product (HVP) via Pearlmutter's method. 

\begin{figure*}[htbp]
    \centering
    \includegraphics[width=\textwidth]{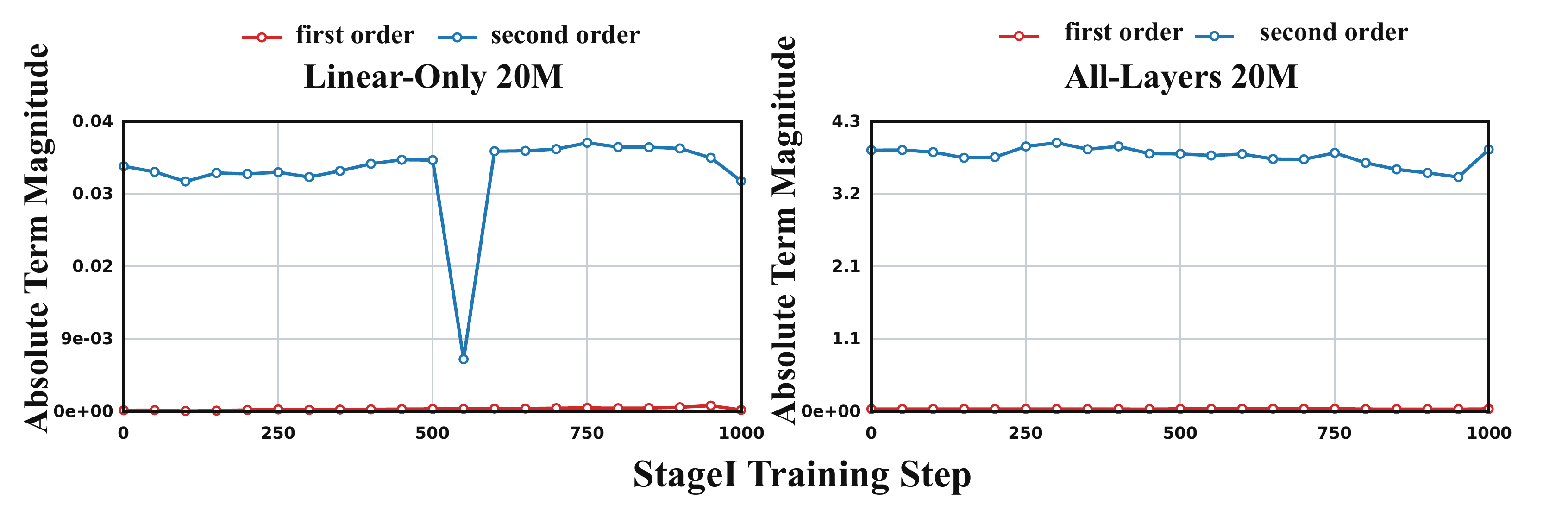}
    \vspace{-6mm}
    \caption{Empirical validation of the Taylor expansion assumption during Stage I training. We plot the absolute magnitudes of the first-order term $|g_s^\top (M_0 \odot \theta)|$ and the second-order term $|\frac{1}{2}(M_0 \odot \theta)^\top H_s (M_0 \odot \theta)|$ over 1000 training steps for both Linear-Only (Left) and All-Layers (Right) variants under a 20M parameter budget. The second-order term dominates the optimization landscape by over two orders of magnitude, validating our theoretical formulation.}
    \label{fig:hvp_validation}
\end{figure*}

Specifically, we evaluate the Linear-Only (20M) and All-Layers (20M) variants on the SD1.5 backbone over 1000 probing steps. As shown in Figure~\ref{fig:hvp_validation}, the magnitude of the first-order term (red line) remains consistently near zero throughout the entire optimization trajectory. In contrast, the second-order term (blue line) dominates the loss variation. Quantitative analysis reveals that the median absolute magnitude of the second-order term is significantly larger than that of the first-order term---approximately $114.6\times$ larger for the Linear-Only variant and $111.2\times$ larger for the All-Layers variant. 

These empirical observations strongly support our theoretical framework: minimizing the probing loss during Stage I is indeed roughly equivalent to minimizing the restricted Hessian quadratic form, thereby effectively discovering a low-interference writable subspace for downstream adaptation.

\section{More Comparison Results on Downstream Dataset Fine-tuning}
\begin{figure*}[t]
    \centering
    \begin{overpic}[width=0.98\textwidth]{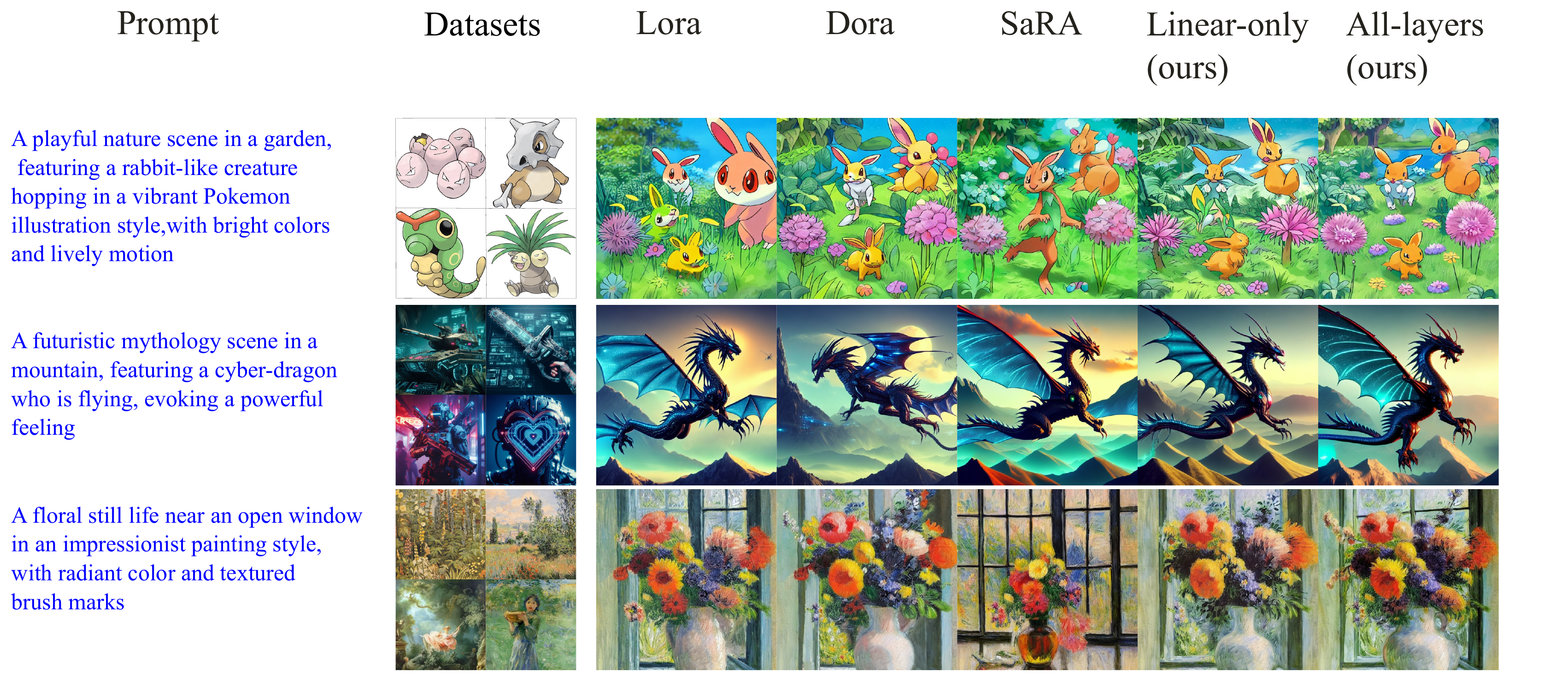}
    \end{overpic}
    \vspace{-3mm}
    \caption{Comparisons of different fine-tuning methods on SD1.5
    }
    \label{fig:comparison2}
\end{figure*}

\begin{figure*}[t]
    \centering
    \begin{overpic}[width=0.98\textwidth]{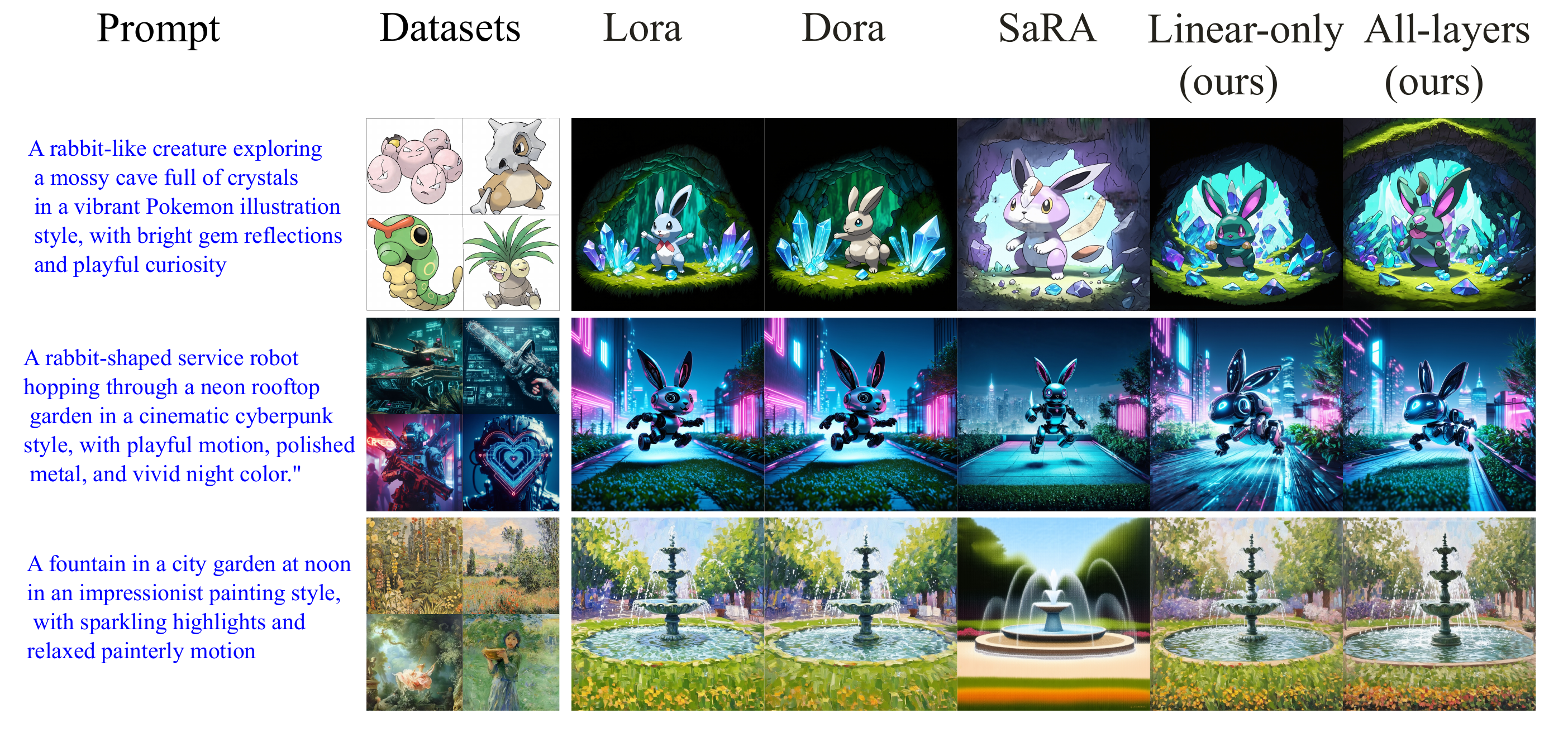}
    \end{overpic}
    \vspace{-3mm}
    \caption{Comparisons of different fine-tuning methods on SD3
    }
    \label{fig:comparison3}
\end{figure*}

In this section, we provide additional comparison results on downstream fine-tuning. 
Figs.~\ref{fig:comparison2} and~\ref{fig:comparison3} show additional comparisons on SD1.5 and SD3, respectively. 
Across different downstream datasets and prompts, our method generally achieves superior generation quality . 

\begin{table*}[!htbp]
\centering
\scriptsize
\setlength{\tabcolsep}{4pt}
\caption{Dense low-budget comparison under the quick protocol on SD1.5. We compare LoRA, DoRA, and Linear-Only (Ours) at the measured budgets of 3M, 4M, 6M, and 8M on \texttt{Pokemon}, \texttt{Cyberpunk}, and \texttt{Anime}. tKID is multiplied by $10^3$, while tCLIP and sCLIP are multiplied by $10^2$. 
This table complements the rigorous main comparison by providing a finer-grained view of the budget--performance trade-off in the low-budget regime.}
\label{tab:sd15_low_budget_tradeoff}
\resizebox{\textwidth}{!}{%
\begin{tabular}{c c c c c c c}
\toprule
\multicolumn{7}{c}{\textbf{Pokemon}} \\
\midrule
Params & Method & tKID$(\times 10^3)\downarrow$ & tCLIP$(\times 10^2)\uparrow$ & sFID$\downarrow$ & sCLIP$(\times 10^2)\uparrow$ & Trade-off$\downarrow$ \\
\midrule
\multirow{3}{*}{3M} & LoRA & 28.97 & 32.88 & 51.97 & 31.36 & 1.24 \\
& DoRA & 28.44 & \textbf{33.06} & 52.59 & \textbf{31.64} & 1.23 \\
& Linear-Only (Ours) & \textbf{23.71} & 32.14 & \textbf{51.27} & 31.30 & \textbf{1.00} \\
\cmidrule(lr){1-7}
\multirow{3}{*}{4M} & LoRA & 26.51 & \textbf{33.04} & 52.60 & 31.39 & 1.18 \\
& DoRA & 24.72 & 32.86 & 51.81 & \textbf{31.56} & 1.08 \\
& Linear-Only (Ours) & \textbf{22.90} & 32.05 & \textbf{51.58} & 31.28 & \textbf{1.00} \\
\cmidrule(lr){1-7}
\multirow{3}{*}{6M} & LoRA & 26.46 & \textbf{32.83} & 52.58 & 31.59 & 1.30 \\
& DoRA & 27.20 & 32.76 & \textbf{51.48} & \textbf{31.60} & 1.31 \\
& Linear-Only (Ours) & \textbf{20.77} & 31.97 & 53.55 & 31.51 & \textbf{1.04} \\
\cmidrule(lr){1-7}
\multirow{3}{*}{8M} & LoRA & 25.64 & 32.80 & 52.12 & 31.41 & 1.30 \\
& DoRA & 24.41 & \textbf{32.89} & \textbf{51.69} & \textbf{31.50} & 1.23 \\
& Linear-Only (Ours) & \textbf{19.87} & 31.71 & 52.76 & 31.41 & \textbf{1.02} \\
\midrule
\multicolumn{7}{c}{\textbf{Cyberpunk}} \\
\midrule
Params & Method & tKID$(\times 10^3)\downarrow$ & tCLIP$(\times 10^2)\uparrow$ & sFID$\downarrow$ & sCLIP$(\times 10^2)\uparrow$ & Trade-off$\downarrow$ \\
\midrule
\multirow{3}{*}{3M} & LoRA & 9.86 & 35.72 & 54.03 & 31.36 & 1.32 \\
& DoRA & 10.21 & 35.72 & \textbf{53.23} & \textbf{31.47} & 1.34 \\
& Linear-Only (Ours) & \textbf{7.61} & \textbf{36.07} & 55.13 & 31.38 & \textbf{1.04} \\
\cmidrule(lr){1-7}
\multirow{3}{*}{4M} & LoRA & 8.50 & 35.91 & \textbf{53.56} & \textbf{31.40} & 1.06 \\
& DoRA & 9.77 & 35.86 & 54.64 & 31.37 & 1.24 \\
& Linear-Only (Ours) & \textbf{8.02} & \textbf{36.20} & 55.61 & 31.29 & \textbf{1.04} \\
\cmidrule(lr){1-7}
\multirow{3}{*}{6M} & LoRA & 9.19 & 35.93 & 53.95 & \textbf{31.39} & 1.08 \\
& DoRA & 9.35 & 35.94 & \textbf{53.92} & 31.37 & 1.10 \\
& Linear-Only (Ours) & \textbf{8.51} & \textbf{36.20} & 55.83 & 31.34 & \textbf{1.03} \\
\cmidrule(lr){1-7}
\multirow{3}{*}{8M} & LoRA & 9.36 & 36.02 & 54.06 & \textbf{31.48} & 1.30 \\
& DoRA & 9.74 & 36.06 & \textbf{53.90} & 31.48 & 1.34 \\
& Linear-Only (Ours) & \textbf{7.24} & \textbf{36.33} & 56.01 & 31.42 & \textbf{1.04} \\
\midrule
\multicolumn{7}{c}{\textbf{Anime}} \\
\midrule
Params & Method & tKID$(\times 10^3)\downarrow$ & tCLIP$(\times 10^2)\uparrow$ & sFID$\downarrow$ & sCLIP$(\times 10^2)\uparrow$ & Trade-off$\downarrow$ \\
\midrule
\multirow{3}{*}{3M} & LoRA & 57.74 & \textbf{24.03} & 52.46 & 31.28 & 1.77 \\
& DoRA & 36.23 & 23.44 & \textbf{51.94} & 31.50 & 1.10 \\
& Linear-Only (Ours) & \textbf{33.03} & 23.47 & 54.49 & \textbf{31.51} & \textbf{1.05} \\
\cmidrule(lr){1-7}
\multirow{3}{*}{4M} & LoRA & 54.73 & 23.80 & 53.14 & 31.29 & 1.81 \\
& DoRA & \textbf{30.50} & \textbf{23.92} & \textbf{52.64} & 31.36 & \textbf{1.00} \\
& Linear-Only (Ours) & 30.94 & 23.76 & 56.04 & \textbf{31.42} & 1.08 \\
\cmidrule(lr){1-7}
\multirow{3}{*}{6M} & LoRA & 41.04 & \textbf{23.75} & \textbf{52.70} & 31.25 & 1.20 \\
& DoRA & 35.42 & 23.71 & 53.31 & 31.28 & \textbf{1.05} \\
& Linear-Only (Ours) & \textbf{34.20} & 23.62 & 58.76 & \textbf{31.39} & 1.11 \\
\cmidrule(lr){1-7}
\multirow{3}{*}{8M} & LoRA & 42.78 & 23.56 & \textbf{53.24} & 31.28 & 1.80 \\
& DoRA & \textbf{23.81} & 23.75 & 53.40 & 31.41 & \textbf{1.00} \\
& Linear-Only (Ours) & 35.24 & \textbf{23.83} & 59.89 & \textbf{31.42} & 1.67 \\
\bottomrule
\end{tabular}%
}
\end{table*}

\section{Backbone Fine-tuning on FFHQ}

\paragraph{Motivation.}
In addition to the style and domain adaptation experiments in the main paper,
we further evaluate our method on FFHQ~\cite{karras2019stylegan}, a real
human-face image distribution. This setting is different from conventional
style transfer: the goal is not to inject a new artistic style, but to refine
the pretrained backbone on a real-image subdomain where the base SD1.5 model
shows relatively weak distribution matching under our evaluation protocol.

\paragraph{Setup.}
We use SD1.5 as the backbone and fine-tune it on FFHQ for 10,000 training steps.
All methods are evaluated with 5,000 generated images. We compare LoRA, DoRA,
SaRA, our Linear-Only variant, and our All-Layers variant under the same 8M
trainable-parameter budget. We report FID, KID, KID standard deviation, CLIP
score, and A-MMD.

\begin{table}[t]
\centering
\caption{
Additional backbone fine-tuning results on FFHQ using SD1.5.
All methods are trained for 10,000 steps and evaluated with 5,000 generated
images under an 8M trainable-parameter budget. KID and KID std are multiplied
by $10^3$, while CLIP is multiplied by $10^2$. Lower FID, KID, KID std, and
A-MMD are better, while higher CLIP is better. The base model is included as a
reference without fine-tuning.
}
\label{tab:supp_ffhq_backbone}
\resizebox{0.7\textwidth}{!}{
\begin{tabular}{c l c c c c c}
\toprule
Params & Method & FID $\downarrow$ & KID$(\times 10^3)\downarrow$ & KID std$(\times 10^3)\downarrow$ & CLIP$(\times 10^2)\uparrow$ & A-MMD $\downarrow$ \\
\midrule
-- & Base & 83.25 & -- & -- & 26.09 & 0.10 \\
\cmidrule(lr){1-7}
\multirow{5}{*}{8M} & LoRA & 33.05 & \textbf{15.63} & 5.35 & \textbf{26.29} & 0.06 \\
& DoRA & 35.28 & 18.08 & 5.77 & 26.16 & 0.06 \\
& SaRA & 37.89 & 28.89 & 8.28 & 26.15 & \textbf{0.04} \\
& Linear-Only & \textbf{31.40} & 16.20 & \textbf{5.10} & 25.94 & 0.06 \\
& All-Layers & 35.94 & 23.61 & 7.25 & 26.20 & 0.06 \\
\bottomrule
\end{tabular}
}
\end{table}

\paragraph{Discussion.}
The results show that fine-tuning on FFHQ substantially improves the
face-domain distribution matching of SD1.5. The base model obtains an FID of
83.25, while all fine-tuned variants reduce FID significantly. Among the
compared methods, our Linear-Only variant achieves the best FID, reducing it to
31.40 under the same 8M trainable-parameter budget. It also obtains a
competitive KID score and the lowest KID standard deviation among the
fine-tuned methods. This suggests that the proposed source-prior-driven writable
support is not limited to stylized target-domain transfer, but can also be useful
for refining a pretrained diffusion backbone on a real-image subdomain where
the base model is weak.

At the same time, the FFHQ setting should be interpreted differently from the
style-transfer experiments in the main paper. Since FFHQ is a natural face-image
distribution rather than a stylized target domain, the goal is not to inject a new
visual style, but to improve the model's fidelity on a specific real-world domain.
Therefore, we regard this experiment as backbone domain capability refinement
rather than conventional style adaptation.

\section{Human Preference Study}
\label{app:human-study}

\paragraph{Protocol.}
To complement the automatic FID/KID/CLIP-based evaluation, we conduct a pilot blinded pairwise human preference study.
The study evaluates the same two aspects as our main objective: target-domain adaptation and source-domain capability retention.
For each trial, participants are shown two anonymized images generated from the same prompt by our method and one baseline method, and choose one of three options: ``A is better'', ``B is better'', or ``Tie / No clear preference''.
Method names, file names, and automatic metric scores are hidden from participants, and the left--right order of the two methods is randomized to reduce position bias.

For the target-adaptation task, participants are asked to select the image that better matches the target domain or target style while preserving prompt content and visual quality.
For the source-retention task, participants are asked to select the image that better follows a general open-domain prompt, looks natural, and avoids irrelevant stylization or artifacts.
Responses are collected from two questionnaire forms and aggregated at the domain--baseline comparison level.

\paragraph{Metrics.}
We report percentage-based preference statistics rather than raw vote counts.
Let $W$, $T$, and $L$ denote the internal win, tie, and loss counts for our method in a comparison group.
The non-tie win rate is defined as
\begin{equation}
\mathrm{WinRate}_{\mathrm{nt}} = \frac{W}{W+L},
\end{equation}
and the tie-adjusted preference rate is defined as
\begin{equation}
\mathrm{Pref}_{\mathrm{adj}} =
\frac{W+0.5T}{W+T+L}.
\end{equation}

\begin{table*}[t]
\centering
\small
\setlength{\tabcolsep}{6pt}
\caption{
Human preference results for target-domain adaptation.
For each trial, participants compare images generated by our method and one baseline method under the same prompt.
Only percentage-based preference statistics are reported.
}
\label{tab:human_target}
\resizebox{0.86\textwidth}{!}{
\begin{tabular}{l l c c}
\toprule
Domain & Comparison & Non-tie win rate (\%) & Tie-adjusted preference (\%) \\
\midrule
Cyberpunk & Ours vs. SaRA & 71.4 & 68.8 \\
Cyberpunk & Ours vs. DoRA & 66.7 & 62.5 \\
Cyberpunk & Ours vs. LoRA & 25.0 & 25.0 \\
WikiArt & Ours vs. DoRA & 75.0 & 75.0 \\
WikiArt & Ours vs. SaRA & 87.5 & 87.5 \\
WikiArt & Ours vs. LoRA & 75.0 & 75.0 \\
Pokemon & Ours vs. SaRA & 85.7 & 81.3 \\
Pokemon & Ours vs. DoRA & 71.4 & 68.8 \\
Pokemon & Ours vs. LoRA & 100.0 & 93.8 \\
\bottomrule
\end{tabular}
}
\end{table*}

\begin{table*}[t]
\centering
\small
\setlength{\tabcolsep}{6pt}
\caption{
Human preference results for source-domain retention.
The source-retention study uses general open-domain prompts rather than target-style prompts.
Only percentage-based preference statistics are reported.
}
\label{tab:human_source}
\resizebox{0.86\textwidth}{!}{
\begin{tabular}{l l c c}
\toprule
Adapted domain & Comparison & Non-tie win rate (\%) & Tie-adjusted preference (\%) \\
\midrule
Anime & Ours vs. LoRA & 50.0 & 50.0 \\
Anime & Ours vs. SaRA & 100.0 & 87.5 \\
Anime & Ours vs. DoRA & 25.0 & 25.0 \\
Cyberpunk & Ours vs. DoRA & 42.9 & 43.8 \\
Cyberpunk & Ours vs. SaRA & 66.7 & 62.5 \\
Cyberpunk & Ours vs. LoRA & 80.0 & 68.8 \\
WikiArt & Ours vs. LoRA & 14.3 & 18.8 \\
WikiArt & Ours vs. DoRA & 37.5 & 37.5 \\
WikiArt & Ours vs. SaRA & 80.0 & 68.8 \\
Pokemon & Ours vs. DoRA & 33.3 & 37.5 \\
Pokemon & Ours vs. SaRA & 100.0 & 87.5 \\
Pokemon & Ours vs. LoRA & 33.3 & 37.5 \\
\bottomrule
\end{tabular}
}
\end{table*}

\paragraph{Discussion.}
For target adaptation, our method obtains an average tie-adjusted preference of 70.8\% across the domain--baseline comparison groups.
This indicates that human raters generally prefer the target-domain images produced by our method over the compared baselines, although Cyberpunk against LoRA remains an exception.
For source retention, the average tie-adjusted preference is 52.1\%, showing a more mixed pattern.
Our method is consistently preferred over SaRA in source-retention comparisons, but not uniformly over LoRA or DoRA.
This is consistent with the main paper's conclusion that the proposed method improves the adaptation--retention trade-off in many budget-constrained regimes, but is not a universal winner on every individual target or source criterion.
Because this is a pilot study, we use it as complementary evidence rather than as a replacement for automatic metrics.

\section{Additional Analysis of SVDiff and WaveFT}
\label{app:svdiff-waveft}

\paragraph{Motivation.}
The main paper compares our method with LoRA, DoRA, and SaRA.
To further position our method against compact and selective diffusion fine-tuning methods, we add comparisons with SVDiff ~\cite{svdiff}and WaveFT~\cite{waveft}.
Both methods are informative baselines, but their strongest operating regimes differ from the budget-controlled selective adaptation setting studied in this paper.

\paragraph{SVDiff.}
SVDiff fine-tunes singular values after SVD decomposition.
Under the original singular-value-only parameterization used in our SD1.5 implementation, the largest trainable parameter count available after enabling all covered SVDiff parameters is about 0.2M.
Scaling it to the 1.6M, 21.4M, or 50M budgets used in our main experiments would require adding extra trainable components or changing the update parameterization, which would no longer be a faithful SVDiff comparison.
We therefore report SVDiff as an ultra-compact baseline rather than a strictly budget-matched baseline.

\paragraph{WaveFT.}
WaveFT is also originally designed for an ultra-compact regime, close to a LoRA-rank-1 equivalent.
In our SD1.5 setting, this corresponds to roughly 0.2M trainable parameters.
For completeness, we additionally test a matched 1.6M WaveFT configuration.
This setting directly matches our lowest SD1.5 budget, but also pushes WaveFT outside its intended ultra-compact regime.

\begin{table*}[t]
\centering
\scriptsize
\setlength{\tabcolsep}{4pt}
\caption{
Additional comparison with SVDiff and WaveFT on SD1.5.
KID and KID std are multiplied by $10^3$, while CLIP is multiplied by $10^2$.
SVDiff is reported under its original ultra-compact singular-value-only parameterization.
WaveFT is evaluated under a matched 1.6M trainable-parameter budget.
}
\label{tab:svdiff_waveft}
\resizebox{\textwidth}{!}{
\begin{tabular}{c l l c c c c c c}
\toprule
Params & Method & Dataset & tKID$(\times 10^3)\downarrow$ & KID std$(\times 10^3)\downarrow$ & tCLIP$(\times 10^2)\uparrow$ & sFID$\downarrow$ & sCLIP$(\times 10^2)\uparrow$ & A-MMD$\downarrow$ \\
\midrule
\multirow{3}{*}{0.2M} & \multirow{3}{*}{SVDiff} & WikiArt & 40.70 & 5.20 & 28.41 & 18.83 & 31.54 & 0.0304 \\
& & Anime & 34.50 & 6.50 & 24.12 & 21.20 & 31.40 & 0.0318 \\
& & Cyberpunk & 8.40 & 2.90 & 35.13 & 20.94 & 31.42 & 0.0389 \\
\cmidrule(lr){1-9}
\multirow{3}{*}{1.6M} & \multirow{3}{*}{WaveFT} & WikiArt & 44.47 & 8.57 & 27.04 & 54.73 & 26.97 & 0.1025 \\
& & Anime & 78.46 & 9.37 & 24.37 & 129.15 & 24.58 & 0.2331 \\
& & Cyberpunk & 11.21 & 3.94 & 35.87 & 53.63 & 27.90 & 0.0970 \\
\bottomrule
\end{tabular}
}
\end{table*}

\paragraph{Discussion.}
SVDiff preserves source-domain capability well and can be competitive in the ultra-compact regime.
However, because its trainable parameter count is not naturally adjustable under the original singular-value-only parameterization, it does not provide a matched-budget comparison across the parameter ranges studied in the main paper.
When a larger but still parameter-efficient budget is allowed, our method provides substantially stronger target adaptation: at 21.4M, Linear-Only obtains tKID scores of 17.15, 22.28, and 7.39 on WikiArt, Anime, and Cyberpunk, respectively, compared with SVDiff's 40.70, 34.50, and 8.40.

The matched-budget WaveFT results further show that scaling WaveFT directly to 1.6M causes severe source-side degradation.
Compared with our 1.6M Linear-Only results, WaveFT has much higher source FID on WikiArt, Anime, and Cyberpunk: 54.73/129.15/53.63 versus 19.20/21.71/22.00.
This indicates strong source-side style contamination after adaptation.
Therefore, SVDiff and WaveFT are useful for understanding compact adaptation regimes, while in this work our source-prior-driven selection is more aligned with the budget-controlled adaptation--retention trade-off considered.
%
%
%
\bibliographystyle{splncs04}
\bibliography{src/reference}

\begin{thebibliography}{10}
\providecommand{\url}[1]{\texttt{#1}}
\providecommand{\urlprefix}{URL }
\providecommand{\doi}[1]{https://doi.org/#1}

\bibitem{natural_gradient}
Amari, S.i.: {Natural Gradient Works Efficiently in Learning}. Neural
  Computation  \textbf{10}(2),  251--276 (1998)

\bibitem{bitfit}
Ben~Zaken, E., Goldberg, Y., Ravfogel, S.: {B}it{F}it: Simple
  parameter-efficient fine-tuning for transformer-based masked language-models.
  In: Proceedings of the 60th Annual Meeting of the Association for
  Computational Linguistics (Volume 2: Short Papers). pp.~1--9. Association for
  Computational Linguistics, Dublin, Ireland (May 2022).
  \doi{10.18653/v1/2022.acl-short.1}

\bibitem{waveft}
Bilican, A., Yilmaz, M.A., Tekalp, A.M., Cinbis, R.G.: Exploring sparsity for
  parameter efficient fine tuning using wavelets. ArXiv
  \textbf{abs/2505.12532} (2025)

\bibitem{kid}
Bińkowski, M., Sutherland, D.J., Arbel, M., Gretton, A.: Demystifying {MMD}
  {GAN}s. In: International Conference on Learning Representations (2018)

\bibitem{pmlr-v235-esser24a}
Esser, P., Kulal, S., Blattmann, A., Entezari, R., M\"{u}ller, J., Saini, H.,
  Levi, Y., Lorenz, D., Sauer, A., Boesel, F., Podell, D., Dockhorn, T.,
  English, Z., Rombach, R.: Scaling rectified flow transformers for
  high-resolution image synthesis. In: Proceedings of the 41st International
  Conference on Machine Learning. vol.~235, pp. 12606--12633 (2024)

\bibitem{rigl}
Evci, U., Gale, T., Menick, J., Castro, P.S., Elsen, E.: Rigging the lottery:
  Making all tickets winners (2021)

\bibitem{fisher_information}
Fisher, R.A.: {Theory of Statistical Estimation}. Mathematical Proceedings of
  the Cambridge Philosophical Society  \textbf{22}(5),  700--725 (1925)

\bibitem{lottery_ticket}
Frankle, J., Carbin, M.: The lottery ticket hypothesis: Finding sparse,
  trainable neural networks. arXiv: Learning  (2018)

\bibitem{textual_inversion}
Gal, R., Alaluf, Y., Atzmon, Y., Patashnik, O., Bermano, A.H., Chechik, G.,
  Cohen-Or, D.: An image is worth one word: Personalizing text-to-image
  generation using textual inversion. In: The Eleventh International Conference
  on Learning Representations (2023)

\bibitem{svdiff}
Han, L., Li, Y., Zhang, H., Milanfar, P., Metaxas, D., Yang, F.: Svdiff:
  Compact parameter space for diffusion fine-tuning (2023)

\bibitem{magnitude_pruning}
Han, S., Pool, J., Tran, J., Dally, W.J.: Learning both weights and connections
  for efficient neural network. In: Neural Information Processing Systems
  (2015)

\bibitem{fid}
Heusel, M., Ramsauer, H., Unterthiner, T., Nessler, B., Hochreiter, S.: Gans
  trained by a two time-scale update rule converge to a local nash equilibrium.
  In: Neural Information Processing Systems (2017)

\bibitem{ddpm}
Ho, J., Jain, A., Abbeel, P.: Denoising diffusion probabilistic models (2020)

\bibitem{adapter}
Houlsby, N., Giurgiu, A., Jastrzebski, S., Morrone, B., de~Laroussilhe, Q.,
  Gesmundo, A., Attariyan, M., Gelly, S.: Parameter-efficient transfer learning
  for nlp (2019)

\bibitem{lora}
Hu, E.J., Shen, Y., Wallis, P., Allen-Zhu, Z., Li, Y., Wang, S., Wang, L.,
  Chen, W.: {LoRA: Low-Rank Adaptation of Large Language Models}. In:
  International Conference on Learning Representations (2022)

\bibitem{sara}
Hu, T., Zhang, J., Yi, R., Huang, H., Wang, Y., Ma, L.: Sa{RA}: High-efficient
  diffusion model fine-tuning with progressive sparse low-rank adaptation. In:
  The Thirteenth International Conference on Learning Representations (2025)

\bibitem{karras2019stylegan}
Karras, T., Laine, S., Aila, T.: A style-based generator architecture for
  generative adversarial networks. 2019 IEEE/CVF Conference on Computer Vision
  and Pattern Recognition (CVPR) pp. 4396--4405 (2018)

\bibitem{imagic}
Kawar, B., Zada, S., Lang, O., Tov, O., Chang, H., Dekel, T., Mosseri, I.,
  Irani, M.: Imagic: Text-based real image editing with diffusion models (2023)

\bibitem{sparselora}
Khaki, S., Li, X., Guo, J., Zhu, L., Plataniotis, K.N., Yazdanbakhsh, A.,
  Keutzer, K., Han, S., Liu, Z.: Sparselo{RA}: Accelerating {LLM} fine-tuning
  with contextual sparsity. In: Forty-second International Conference on
  Machine Learning (2025)

\bibitem{ewc}
Kirkpatrick, J., Pascanu, R., Rabinowitz, N., Veness, J., Desjardins, G., Rusu,
  A.A., Milan, K., Quan, J., Ramalho, T., Grabska-Barwinska, A., Hassabis, D.,
  Clopath, C., Kumaran, D., Hadsell, R.: Overcoming catastrophic forgetting in
  neural networks. Proceedings of the National Academy of Sciences
  \textbf{114}(13),  3521–3526 (Mar 2017). \doi{10.1073/pnas.1611835114}

\bibitem{vera}
Kopiczko, D.J., Blankevoort, T., Asano, Y.M.: Ve{RA}: Vector-based random
  matrix adaptation. In: The Twelfth International Conference on Learning
  Representations (2024)

\bibitem{custom_diffusion}
Kumari, N., Zhang, B., Zhang, R., Shechtman, E., Zhu, J.Y.: Multi-concept
  customization of text-to-image diffusion (2023)

\bibitem{prompt_tuning}
Lester, B., Al-Rfou, R., Constant, N.: The power of scale for
  parameter-efficient prompt tuning. In: Proceedings of the 2021 Conference on
  Empirical Methods in Natural Language Processing. pp. 3045--3059. Association
  for Computational Linguistics (Nov 2021).
  \doi{10.18653/v1/2021.emnlp-main.243}

\bibitem{prefix_tuning}
Li, X.L., Liang, P.: Prefix-tuning: Optimizing continuous prompts for
  generation (2021)

\bibitem{shadowpeft}
Li, X., Li, Z., fung Andrew~Lee, T., Li, J., Xie, H., Li, Q.: Shadowpeft:
  Shadow network for parameter-efficient fine-tuning (2026)

\bibitem{lwf}
Li, Z., Hoiem, D.: Learning without forgetting. IEEE Transactions on Pattern
  Analysis and Machine Intelligence  \textbf{40},  2935--2947 (2016)

\bibitem{dora}
yang Liu, S., Wang, C.Y., Yin, H., Molchanov, P., Wang, Y.C.F., Cheng, K.T.,
  Chen, M.H.: Do{RA}: Weight-decomposed low-rank adaptation. In: Forty-first
  International Conference on Machine Learning (2024)

\bibitem{gem}
Lopez-Paz, D., Ranzato, M.: Gradient episodic memory for continual learning.
  In: Neural Information Processing Systems (2017)

\bibitem{tspeft}
Ma, D., Dai, Z., Xin, Z., Wang, S., Yang, J., Fei, H.: Ts-peft: Unveiling
  token-level redundancy in parameter-efficient fine-tuning (2026)

\bibitem{compacter}
mahabadi, R.K., Henderson, J., Ruder, S.: Compacter: Efficient low-rank
  hypercomplex adapter layers. In: Beygelzimer, A., Dauphin, Y., Liang, P.,
  Vaughan, J.W. (eds.) Advances in Neural Information Processing Systems (2021)

\bibitem{packnet}
Mallya, A., Lazebnik, S.: Packnet: Adding multiple tasks to a single network by
  iterative pruning. 2018 IEEE/CVF Conference on Computer Vision and Pattern
  Recognition pp. 7765--7773 (2017)

\bibitem{kfac}
Martens, J., Grosse, R.: Optimizing neural networks with kronecker-factored
  approximate curvature (2020)

\bibitem{catastrophic_forgetting}
McCloskey, M., Cohen, N.J.: Catastrophic interference in connectionist
  networks: The sequential learning problem. Psychology of Learning and
  Motivation  \textbf{24},  109--165 (1989)

\bibitem{continual_learning_survey}
Parisi, G.I., Kemker, R., Part, J.L., Kanan, C., Wermter, S.: Continual
  lifelong learning with neural networks: A review. Neural networks : the
  official journal of the International Neural Network Society  \textbf{113},
  54--71 (2018)

\bibitem{clip}
Radford, A., Kim, J.W., Hallacy, C., Ramesh, A., Goh, G., Agarwal, S., Sastry,
  G., Askell, A., Mishkin, P., Clark, J., Krueger, G., Sutskever, I.: Learning
  transferable visual models from natural language supervision. In:
  International Conference on Machine Learning (2021)

\bibitem{stable_diffusion}
Rombach, R., Blattmann, A., Lorenz, D., Esser, P., Ommer, B.: {Stable
  Diffusion}. \url{https://github.com/CompVis/stable-diffusion} (2022)

\bibitem{ldm}
Rombach, R., Blattmann, A., Lorenz, D., Esser, P., Ommer, B.: High-resolution
  image synthesis with latent diffusion models (2022)

\bibitem{dreambooth}
Ruiz, N., Li, Y., Jampani, V., Pritch, Y., Rubinstein, M., Aberman, K.:
  Dreambooth: Fine tuning text-to-image diffusion models for subject-driven
  generation. 2023 IEEE/CVF Conference on Computer Vision and Pattern
  Recognition (CVPR) pp. 22500--22510 (2022)

\bibitem{ddim}
Song, J., Meng, C., Ermon, S.: Denoising diffusion implicit models. In:
  International Conference on Learning Representations (2021)

\bibitem{score_sde}
Song, Y., Sohl-Dickstein, J., Kingma, D.P., Kumar, A., Ermon, S., Poole, B.:
  Score-based generative modeling through stochastic differential equations
  (2021)

\bibitem{fishmask}
Sung, Y.L., Nair, V., Raffel, C.: Training neural networks with fixed sparse
  masks. ArXiv  \textbf{abs/2111.09839} (2021)

\bibitem{relora}
Tian, C., Wei, X., Liu, H., Luo, Y., Guo, Z., Li, L.: Less is more:
  Resource-efficient low-rank adaptation (2025)

\bibitem{kasa}
Wang, F., Jiang, J., Park, C., Kim, S., Tang, J.: Ka{SA}: Knowledge-aware
  singular-value adaptation of large language models. In: The Thirteenth
  International Conference on Learning Representations (2025)

\bibitem{moslora}
Wu, T., Wang, J., Zhao, Z., Wong, N.: Mixture-of-subspaces in low-rank
  adaptation. In: Proceedings of the 2024 Conference on Empirical Methods in
  Natural Language Processing. pp. 7880--7899. Association for Computational
  Linguistics, Miami, Florida, USA (Nov 2024).
  \doi{10.18653/v1/2024.emnlp-main.450}

\bibitem{mixture_space_peft}
Zhang, B., Tao, J., Zilang, Z., Maatouk, A., Yang, M., Ying, R.:
  Parameter-efficient fine-tuning of {LLM}s with mixture of space experts
  (2026)

\bibitem{controlnet}
Zhang, L., Rao, A., Agrawala, M.: Adding conditional control to text-to-image
  diffusion models (2023)

\bibitem{adalora}
Zhang, Q., Chen, M., Bukharin, A., He, P., Cheng, Y., Chen, W., Zhao, T.:
  Adaptive budget allocation for parameter-efficient fine-tuning. ArXiv
  \textbf{abs/2303.10512} (2023)

\bibitem{sparse_training_survey}
Zhang, Z., Xu, Y., Yang, J., Li, X., Zhang, D.: A survey of sparse
  representation: Algorithms and applications. IEEE Access  \textbf{3},
  490–530 (2015). \doi{10.1109/access.2015.2430359}

\bibitem{flylora}
Zou, H., Zang, Y., Xu, W., Zhu, Y., Ji, X.: Flylora: Boosting task decoupling
  and parameter efficiency via implicit rank-wise mixture-of-experts. ArXiv
  \textbf{abs/2510.08396} (2025)

\end{thebibliography}

%




\end{document}